\documentclass[preprint,12pt]{elsarticle}

\usepackage{mathtools, amsmath, amssymb, amsthm}
\usepackage{layouts}
\usepackage{hyperref}
\usepackage{float}
\usepackage{comment}
\usepackage{graphicx}
\usepackage[usenames,dvipsnames]{xcolor}
\usepackage{algorithm, algpseudocode, algorithmicx}
\usepackage{hyperref}
\usepackage{multirow}
\usepackage{booktabs}
\usepackage{bm}
\usepackage{bbm}
\usepackage{subcaption}

\usepackage{adjustbox}

\makeatletter
\def\ps@pprintTitle{%
  \let\@oddhead\@empty
  \let\@evenhead\@empty
  \let\@oddfoot\@empty
  \let\@evenfoot\@oddfoot
}
\makeatother

\newtheorem{proposition}{Proposition}
\newtheorem{definition}{Definition}

\newtheorem{lemma}{Lemma}

\let\vec\mathbf

\def\I{{\mathbb{I}}}

\def\E{{\mathbb E}}

\def\R{{\mathbb R}}
\def\C{{\mathbb C}}

\def\K{{\vec{K}}}
\def\X{{\vec{X}}}

\def\I{{\vec{I}}}

\newcommand{\z}{\bm{{z}}}
\newcommand{\sz}{{z}}
\newcommand{\x}{\bm{{x}}}
\newcommand{\y}{\bm{{y}}}
\newcommand{\sx}{{{x}}}

\def\sX{{\mathcal{X}}}
\def\sY{{\mathcal{Y}}}
\def\sH{{\mathcal{H}}}
\def\sF{{\mathcal{F}}}

\def\t{{\intercal}}

\DeclareMathOperator*{\argmin}{arg\,min}

\DeclareMathOperator{\prox}{prox}
\DeclareMathOperator*{\minimize}{minimize}

\let\norm\undefined 
\DeclarePairedDelimiter\norm{\lVert}{\rVert}

\renewcommand{\epsilon}{\varepsilon}

\newcommand{\rffnet}{{\texttt{RFFNet}}}

\journal{Information Sciences}

\begin{document}

\begin{frontmatter}



\title{\texttt{RFFNet}: Large-Scale Interpretable Kernel Methods via Random Fourier Features}

\author[a,b]{Mateus P. Otto}
\author[a]{Rafael Izbicki}

\affiliation[a]{organization={Department of Statistics, Federal University of São Carlos}, city = {São Carlos}, state = {São Paulo}, country = {Brazil}}
\affiliation[b]{organization={Instituto de  Ciências Matemáticas e de Computação, Universidade de São Paulo}, city = {São Carlos}, state = {São Paulo}, country = {Brazil}}

\begin{abstract}

Kernel methods provide a flexible and theoretically grounded approach to nonlinear and nonparametric learning. While memory and run-time requirements hinder their applicability to large datasets, many low-rank kernel approximations, such as random Fourier features, were recently developed to scale up such kernel methods. However, these scalable approaches are based on approximations of isotropic kernels, which cannot remove the influence of irrelevant features. In this work, we design random Fourier features for a family of automatic relevance determination (ARD) kernels, and introduce \rffnet{}, a new large-scale kernel method that learns the kernel relevances' on the fly via first-order stochastic optimization. We present an effective initialization scheme for the method's non-convex objective function, evaluate if hard-thresholding \rffnet{}'s learned relevances yield a sensible rule for variable selection, and perform an extensive ablation study of \rffnet{}'s components.
Numerical validation on simulated and real-world data shows that our approach has a small memory footprint and run-time, achieves low prediction error, and effectively identifies relevant features, thus leading to more interpretable solutions. We supply users with an efficient, PyTorch-based library, that adheres to the \textit{scikit-learn} standard API and code for fully reproducing our results.
\end{abstract}



\begin{keyword}
supervised learning \sep kernel methods \sep feature importance


\end{keyword}

\end{frontmatter}

\section{Introduction}
\label{sec:introduction}

Statistical learning methods based on kernels are successfully applied in many fields, including biology \citep{schoelkopf2004kernel}, social sciences \citep{hainmueller2014kernel}, physics \citep{paine2023quantum}, and astronomy \citep{foremanmackey2017fast, wachman2009kernels}.  Given their nonparametric nature and many theoretical guarantees, kernel methods allow for principled modeling of complex relationships in real-world data.
These methods have also sparked extensive methodological research in statistics, leading to the development of kernel-based tools for feature selection \citep{he2021efficient, jordan2021self}, causal inference \citep{zhang2021instrument}, hypothesis testing \citep{jitkrittum2020testing, liu2021meta, shekhar2022permutation, scetbon2022asymptotic}, and privacy \citep{balog2018differentially}. In particular, cornerstones of machine learning, such as kernel ridge regression, Gaussian processes \citep{rasmussen2006gaussian}, kernel principal component analysis \citep{schoelkopf2002learning}, support vector machines \citep{cortes1995support}, and neural tangent kernels \citep{jacot2018neural} are based on kernels. 

Nevertheless, the applicability of kernel methods is limited by two main factors: scalability and interpretability. Firstly, since these methods operate on all pairs of observations, they require computation (e.g. matrix inversions) and storage that scales at least quadratically in the sample size, which can be prohibitive for large datasets. Fortunately, several kernel approximations were recently developed for scaling up kernel methods, such as random Fourier features methods (RFF) \citep{rahimi2007random,le2013fastfood, curto2020mckernel} and the Nyström approximation \citep{rudi2017falkon, gittens2013revisiting}. 

Secondly, because kernel methods implicitly embed the original feature space into a high-dimensional ambient space, they often lose information about which features of input data are relevant for prediction \citep{chen2023kernel}. This, in turn, translates into an interpretability shortage and also leads to poor predictive performance in problems with
many irrelevant variables \citep{lafferty2008rodeo, bertin2008selection}. This shortage is particularly significant considering the widespread presence of kernel methods in the toolsets of statistics practitioners. 

\subsection{Method Overview and Novelty}

\begin{figure}[!hbt]
    \centering
    \includegraphics{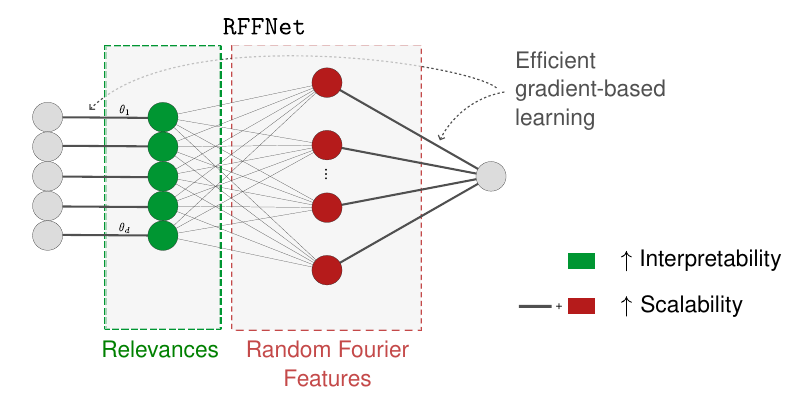}
    \caption{\rffnet{} combines fast kernel approximations to ARD kernels to carefully initialized and efficient stochastic gradient descent methods. The core part of \rffnet{} is implemented as a single PyTorch layer and can be seamlessly connected to other PyTorch-based models, making \rffnet{} highly modular.}
    \label{fig:rffnet-diagram}
\end{figure}

To address the interpretability issue, a common approach is to introduce feature weighting in the kernel parametrization. This process generates the so-called automatic relevance determination (ARD) kernels \citep{rasmussen2006gaussian}, as the ARD Gaussian kernel 
\begin{equation*}
    k_\theta(\x, \x') = \exp\left[-\frac{1}{2} \sum_{j=1}^d \theta_j^2 (x_j-x'_j)^2\right],
\end{equation*}
where $x_j$ is the $j$-th component of the feature vector $\x$ and $\theta_j$ is the relevance of the $j$-th feature. In practice, it is expected that if the relevance vector $\theta$ is estimated using available data, then $\theta_i$ would be shrunk to zero if $x_i$ corresponds to an irrelevant feature.

This work introduces \rffnet{} (see Figure~\ref{fig:rffnet-diagram}), a novel approach for large-scale, interpretable kernel methods. Our method relies on three key components:
\begin{enumerate}
    \item \textbf{Fast kernel approximation via random Fourier features:} This approach significantly reduces the number of parameters to be estimated and decreases the computational overhead of using kernels.

    \item \textbf{Well-designed automatic relevance determination kernels:} We demonstrate that ARD kernels
    correspond to spectral densities wherein kernel relevances act as scale parameters. We then show that it is possible to decouple $\theta$ from the random features map that approximates an ARD kernel's evaluations and promptly estimate $\theta$ with data. 

    \item \textbf{Efficient first-order stochastic optimization methods:} carefully initialized parameters and mini-batching help coping with \rffnet{}'s nonconvex objective function.
\end{enumerate}

In practice,  given a training sample $\{(\x_i, y_i) \}_{i=1}^n$ and an ARD kernel $k_\theta$, \rffnet{} is designed to mimic the solution to the following problem:  find the relevances $\theta$ and the prediction function $g$  that
\begin{equation}
    \minimize_{\theta \in \R^{p}_+, g \in \sH_\theta} \frac{1}{n } \sum_{i=1}^n \ell \left( y_i, g(\x_i) \right) + \lambda \norm{g}_{\sH_\theta}^2,
    \label{eq:original-prob}
\end{equation}
where $\sH_\theta$ is the reproducing kernel Hilbert space (RKHS) associated with $k_\theta$. In its general form, ~\eqref{eq:original-prob} is a known problem in feature selection for kernel methods \citep{chen2017kernel, jordan2021self}. However, as we will show, the restriction to ARD kernels makes~\eqref{eq:original-prob} particularly well-suited to random features approximations. In addition, possible drawbacks of optimizing~\eqref{eq:original-prob} can be tackled with modern stochastic first-order methods \citep{kingma2014adam}. This is the direction taken by \rffnet{}.

\subsection{Relation to other work}\label{sec:related-work}

There is an extensive literature on kernel approximations and its applications to machine learning \citep{rahimi2007random, rahimi2008weighted, le2013fastfood, curto2020mckernel, han2022random, kar2012random, ma2017diving, rudi2017falkon, yang2015la, lazarogredilla2010sparse, quinonerocandela2005unifying}. However, much of the focus has been on approximating isotropic kernels, which weigh relevant and irrelevant features equally in the kernel. 

There are few exceptions to this trend. A La Carte \citep{yang2015la} leverages on Fastfood \citep{le2013fastfood} to learn the spectrum of expressive kernels via marginal likelihood optimization. It is similar to Sparse Spectrum Gaussian Process Regression \citep{lazarogredilla2010sparse}, which uses the inducing points approximation \citep{quinonerocandela2005unifying} instead. These methods significantly differ from \rffnet{}, since they: (i) require solving two separate problems, the marginal likelihood optimization and the kernel ridge matrix inversion, and (ii) are restricted to the regression setting.
Falkon \citep{rudi2017falkon} is more akin to our approach. It is based on the Nyström approximation to the kernel matrix and uses first-order optimization to solve kernel ridge regression problems. Falkon \citep{meanti2022efficient} also supports optimizing kernel hyperparameters, such as the relevances of an ARD kernel, but this requires minimizing an additional objective.
Nevertheless, interpretability is rarely a focal point in these approaches, which prioritize low memory usage and run-time. 

Our approach is closer in spirit to Sparse Random Fourier Features \citep{gregorova2018large}, which uses random Fourier features approximations for ARD kernels, tailoring the objective function and optimization constraints to induce feature selection. Still, significant differences set us apart. First, \rffnet{} handles any differentiable loss function, while SRFF is specifically designed for squared error loss problems in regression. Second, \rffnet{} is based on jointly minimizing the kernel relevances and remaining parameters with efficient stochastic gradient descent methods. In fact, using mini-batches is an indispensable component of \rffnet{}. SRFF, instead, is based on alternately minimizing two objectives, a procedure that can cause indefinite cycling of the algorithm \citep{powell1973search} and often leads to poor performance. Third, the relevances output by SRFF after a single run are unreliable sources 
of feature importances. Finally, \rffnet{} is implemented as an efficient library and can be promptly integrated into data analysis pipelines based on the widely adopted \textit{scikit-learn} standard.

Our work also overlaps with the literature in kernel learning \citep{zien2007multiclass, sinha2016learning, lu2016large, li2020automated, lanckriet2004learning, kloft2011lp, goenen2011multiple}. Nonetheless, kernel learning is often based on alignment objectives, which seek to produce a meaningful representation of data using kernels \citep{sinha2016learning}. In this scenario, even after learning the kernel, users would need to train a kernel-based predictive model to produce outputs for given labeled data. Within the literature of kernel learning, Automated Spectral Kernel Learning (ASKL) \citep{li2020automated} has the most resemblance with \rffnet{}. However, ASKL does not update the lengthscales of the kernel it seeks to learn and thus falls short of removing the influence of irrelevant features. Consequently, it is not possible to subsidize interpretative claims based on ASKL-learned spectral density frequencies.

Finally, other approaches to remove irrelevant features in kernel methods include cross-validation \citep{keerthi2006efficient} and recursive feature backward elimination \citep{guyon2002gene, louw2006variable}. These methods, however, do not scale with the number of features and the sample size. An alternative approach more related to our method
is to develop a loss function that includes feature-wise relevances as part of the objective function. This is done by 
\citet{allen2013automatic}, which proposes an
iterative feature extraction method. Unfortunately, since this procedure does not use kernel approximations, it has prohibitive memory and run-time requirements.

Recently, \citet{jordan2021self} described a new sparsity-inducing mechanism for kernel methods. Although the mechanism is based on optimizing a vector of weights that enters the objective as a data scaling, similar to~\eqref{eq:original-prob}, the solution is specific to kernel ridge regression and metric learning. In addition, it does not employ any kernel matrix approximation and thus has limited applicability to large datasets.

\subsection{Notation and Organization}

For any integer $n \ge 1$, we denote by $[n]$ the set $\{ 1, 2, \dots, n\}$. In addition, we indicate by $\circ$ the element-wise vector multiplication; that is, if $\x, \y \in \R^d$, then $\x \circ \y = (x_1 y_1, \dots, x_d y_d)$. We write $I_d$ for the $d \times d$ identity matrix and $\mathbf{1}_d = (1, \dots, 1)$ for the $d$-dimensional vector of ones. If $z \in \C$, then $z^*$ is its complex conjugate.

The remainder of the paper is organized as follows. We give an overview of kernel methods, random Fourier features, and automatic relevance determination (ARD) kernels in Section~\ref{sec:background}. In Section~\ref{sec:rffnet}, we introduce our method and its properties. In Section~\ref{sec:applications} we validate our approach in simulated and real-world datasets. Finally, in Section~\ref{sec:final}, we conclude with a brief discussion and directions for future work. All proofs are deferred to~\ref{appendix:proofs}.

\section{Background}
\label{sec:background}

\subsection{Kernel methods}
\label{subsec:kernelmethods}

We follow the standard formalization of the supervised learning setting \citep{shalevshwartz2014understanding, mohri2018foundations}. Let $\sX \subseteq \R^d$ be the instance space, $\sY$ a label space, and $\rho(x, y)$ a probability measure on $\sX \times \sY$. Throughout this paper, we assume $\sY \subseteq \mathbb{R}$ for regression tasks and $\sY = \{ 0, 1\}$ for binary classification tasks. We denote by  $\ell: \sY \times \sY \to \R^+ \cup \{ + \infty \}$ a loss function (e.g., squared error loss, cross-entropy). Given a finite training sample $\{ (\x_i, y_i) \}_{i=1}^n$ sampled independently from the joint measure $\rho$, the goal of supervised learning is to find a hypothesis $f: \sX \to \sY$ with small expected loss or generalization error $\mathcal{R}(f) = \E[ \ell(f(\X), Y)]$, where the expectation is taken with respect to the pair $(\X, Y) \sim \rho$.

In kernel methods, we restrict $f$ to the Reproducing Kernel Hilbert Space (RKHS) $\sH$ uniquely associated to a positive semi-definite (PSD) kernel $k: \sX \times \sX \to \R$. 
We will consider kernel methods that result from solving the broad class of regularized loss problems on $\sH$ given by
\begin{equation}
    f^* \in \argmin_{f \in \sH} \frac{1}{n} \sum_{i=1}^n \ell(y_i, f(\x_i))  + \lambda ||f||^2_{\sH},
    \label{eq:rkhs-opt}
\end{equation}
where $\lambda > 0$ is the regularization hyperparameter, and $\norm{\cdot}_{\sH}$ is the RKHS norm (see, for instance, \citet{wainwright2019high} for a explicit characterization). 

The representer theorem \citep{kimeldorf1971some, mohri2018foundations} guarantees that the problem in~\eqref{eq:rkhs-opt} has a solution of the form $f^*(\cdot) = \sum_{i=1}^n \alpha^*_i k(\x_i, \cdot)$ with $\alpha^* \in \R^n$. It can be readily shown that $\alpha^*$ satisfies
\begin{equation}
 \alpha^* \in \argmin_{\alpha \in \R^n}  \frac{1}{n} \sum_{i=1}^n \ell(y_i, (\mathbf{K}\alpha)_i)  + \lambda \alpha^\t \K \alpha,
 \label{eq:rkhs-opt-coef} 
\end{equation}
where $\K$ is the kernel matrix, with elements $\K_{ij} = k(\x_i, \x_j)$, and $(\mathbf{K}\alpha)_i = \sum_{j=1}^n \alpha_j k(\x_i, \x_j)$.

If we choose $\ell(y, y') = (y - y')^2$,
the squared error loss, then problem \eqref{eq:rkhs-opt-coef} is known as kernel ridge regression (KRR). In this case, the solution to \eqref{eq:rkhs-opt-coef} is $\alpha^* = (\K + n \lambda \I)^{-1} \y$, where $\y = (y_1, \dots, y_n)^\t$. Hence the scalability problem: computing the KRR estimator requires $\Theta(n^3)$ time and $\Theta(n^2)$ space \citep{avron2017random}, which can be prohibitive for large datasets. In general, computing a solution to \eqref{eq:rkhs-opt-coef} has, at least, $O(n^2)$ time and space complexities \citep{rudi2017falkon}.

To reduce this computational cost, it is usual to resort to kernel matrix approximations \citep{rahimi2007random, rudi2017generalization, le2013fastfood, rudi2017falkon}
that can effectively scale kernel methods to large datasets, as discussed in Section~\ref{sec:related-work}.

\subsection{Random Fourier Features}
\label{subsec:rff}

Random Fourier features \citep{rahimi2007random, rahimi2008weighted} is a widely used and theoretically grounded \citep{rudi2017generalization, li2021towards} framework for scaling up kernel methods. Its guiding idea is to construct a feature map that approximates kernel evaluations of a given shift-invariant PSD kernel.

Let $k: \sX \times \sX \to \R$ be a continuous, PSD, and shift-invariant kernel of the form $k(\x, \x') = G(\x-\x')$. Then, by Bochner's Theorem\footnote{Here, since $\sX \subseteq \R^d$, the domain is a locally compact set.} \citep{mohri2018foundations}, $k$ is the Fourier transform of a bounded non-negative measure $p(\cdot)$. That is,
\begin{equation*}
   G(\x) = \int_{\sX} p(\omega) e^{i \omega^\t \x} d \omega.
\end{equation*}
If we further assume that $G(0) = 1$, then $p(\cdot)$ is a probability density called the spectral density of the kernel. In this case, the kernel function can be written as
\begin{align}
    k(\x, \x') = G(\x-\x') &= \int_{\sX} e^{ i \omega^\t (\x-\x') } p(\omega) d \omega  \nonumber \\
    &= \int_{\sX} \phi(\x) \phi^*(\x') p(\omega) d \omega \nonumber \\ 
    &= \E_{\omega} [ \phi(\x) \phi^*(\x')], \label{eq:rff-exp}
\end{align}
where $\phi(\x)$ is defined as $\phi(\x) = e^{i \omega^\t \x}$. Since we assumed $k$ is real-valued, then~\eqref{eq:rff-exp} can be further simplified as
\begin{align*}
    k(\x, \x') &= \E_{\omega} [ \cos (\omega^\t (\x-\x'))]. \label{eq:rff-exp-real}
\end{align*}
From this point, the construction of random Fourier features proceeds as follows \citep{rahimi2007random}: sample $\{\omega_i\}_{i=1}^s \sim p(\omega)$ independently and $\{ b_i \}_{i=1}^{s} \sim \mathsf{Uniform}(0, 2\pi)$, then the feature map $\z:  \R^d \to \R^s$ with entries
\begin{equation}
    z_i(\x) = \sqrt{2/s} \cos(\omega_i^\t \x + b_i)
    \label{eq:rff-entries}
\end{equation} satisfies 
\begin{equation*}
    \E_{\omega, b}[ \z(\x)^\t \z(\x')] = \E_{\omega} [ \cos (\omega^\t (\x-\x'))] =  k(\x, \x').
\end{equation*}
In other words, the Monte-Carlo estimate
\begin{equation}
    \hat{k}(\x, \x') = \z(\x)^\t \z(\x') = \frac{1}{s} \sum_{i=1}^s \sz_i(\x) \sz_i(\x')
    \label{eq:approx-kernel}
\end{equation}
is an unbiased estimator of $k(\x, \x')$, which can be used to approximate kernel evaluations. 

However, the benefits of learning with random Fourier features are not fully evident when working at the level of kernel matrix approximations. First, observe that the function $\hat{k}: \sX \times \sX \to \R$ defined in~\eqref{eq:approx-kernel} is itself a PSD kernel \citep{wainwright2019high}, and thus, it is uniquely associated to a reproducing kernel Hilbert space $\widehat{\sH}$ (not necessarily contained in $\sH$). Thus, we can also solve problem~\eqref{eq:rkhs-opt} over $\widehat{\sH}$ \citep{li2021towards}. In this case, by the representer theorem, the optimal solution has the form
\begin{equation}
    \widehat{f}^*(\x) = \sum_{i=1}^s \beta^*_i \z_i(\x) = \z(\x)^\t \beta^*,
    \label{eq:rff-approx}
\end{equation}
where $\beta^* \in \R^s$ satisfies
\begin{equation}
   \beta^* \in \argmin_{\beta \in \R^s}  \frac{1}{n} \sum_{i=1}^n \ell(y_i, \z(\x_i)^\t \beta )  + \lambda \norm{\beta}^2.
   \label{eq:approx-problem}
\end{equation}
This is an optimization problem in $\R^s$ instead of $\R^n$ and can be solved much more efficiently than their full kernel matrix counterpart in~\eqref{eq:rkhs-opt-coef}. For instance, if $\ell$ is chosen as the squared error loss, then problem \eqref{eq:approx-problem} can be written as
\begin{equation}
   \beta^* = \argmin_{\beta \in \R^s}  \frac{1}{n} \norm{\y - \mathbf{Z} \beta}_2^2  + \lambda \norm{\beta}^2,
   \label{eq:approx-problem-krr}
\end{equation}
where $\mathbf{Z} \in \R^{n \times s}$ is defined as the matrix which has its $i$-th row equal to $\z(x_i)^\t$. The solution to~\eqref{eq:approx-problem-krr} is $\beta^* = (\mathbf{Z}^\t \mathbf{Z} + n \lambda \mathbf{I}_s)^{-1} \mathbf{Z}^\t \y$, which can be computed in $O(ns^2)$ time and $O(ns)$ storage, rendering random Fourier features much attractive if $s \ll n$. A fundamental question is whether meeting this requirement on $s$ compromises the statistical quality of the resulting kernel estimator. Several papers have answered this question in the negative \citep{li2021towards, rudi2017generalization, bach2017equivalence, avron2017random}, demonstrating that RFF-based estimators asymptotically (in the sample size) match the optimality of full-kernel estimators in the $s \ll n$ regime. Despite these results, except for \citet{yao2023error}, existing literature does not offer guidance on numerically determining the optimal $s$, creating a gap between theory and practice. In our approach, \rffnet{}, users are encouraged to adjust the number of random features until their expected accuracy is achieved. Crucially, our experiments show that even a small, fixed
number of random features gives satisfactory results.

\subsection{Automatic Relevance Determination (ARD) kernels}
\label{subsec:ard}

Automatic Relevance Determination (ARD) kernels \citep{neal1996bayesian} are widely used for variable selection in Bayesian regression (e.g., Gaussian processes \citep{rasmussen2006gaussian, dance2022fast}, sparse Bayesian learning \citep{tipping2001sparse}), support vector machines \citep{keerthi2006efficient, grandvalet2002adaptive, allen2013automatic}, and self-penalizing objectives \citep{jordan2021self, ruan2021taming}. This family of kernels, which includes the usual Gaussian, Laplace, Cauchy, and Matérn kernels, is generated by introducing continuous feature weights in shift-invariant kernels, adding a layer of interpretability atop them by controlling how features contribute to the kernel value.

\begin{definition}
A function $k_{\theta} : \sX \times \sX \to \R$ is an automatic relevance determination (ARD) kernel if it is a continuous, positive-semidefinite, and shift-invariant kernel of the form
\begin{equation}
    k_\theta (\x, \x') = G[ \boldsymbol{}\theta \circ (\x-\x')],
    \label{eq:ard}
\end{equation}
and is uniquely parameterized by $\theta \in \R^d_+$, the relevance vector.   
\end{definition}

ARD kernels are naturally interpretable. Take $\sH_\theta$ as the RKHS associated with an ARD kernel $k_\theta$. Each entry $\theta_i$ of the relevance vector is the reciprocal of the lengthscale of variation of the $i$-th coordinate of functions $f \in \sH_{\theta}$; that is, the smaller the $\theta_i$, the slower the functions of $\sH$ will change in the direction of the $i$-th feature. In the extreme case $\theta_i = 0$, these functions will not depend on the $i$-th feature. Hence, removing the influence of a feature requires shrinking to zero the corresponding relevance parameter.

In the Bayesian learning setting, this shrinkage is accomplished by giving $\theta$ a hyper-prior and maximizing the Type-II marginal likelihood with respect to $\theta$  \citep{rasmussen2006gaussian, dance2022fast}, which naturally leads to sparse $\theta$.
In contrast, within the empirical risk minimization (ERM) setting, filtering out the irrelevant features might improve the method's predictive performance by aligning the data representation, embodied by the kernel, with the data itself \cite{goenen2011multiple, chen2023kernel}. Therefore, including $\theta$ as an additional parameter in the empirical risk minimization should sparsify $\theta$ and underscore pertinent features to the given predictive task. Crucially, in the context of ERM, the accuracy of predictions indicates how trustworthy the interpretation is: solutions with lower errors are likely to match the data better and how it's represented. As a result, striving for the lowest prediction error and reliable interpretation are naturally linked objectives in this scenario.

Nevertheless, the definition of ARD kernels in~\eqref{eq:ard} does not rule out dependence on terms involving more than one feature. For instance, the kernel $k_{\theta}: \R^2 \times \R^2 \to \R$ with law
\begin{equation*}
    k_\theta(\x, \x') = e^{-\bigl[\theta_1^2 |x_1-x_1'|^2 - 2 \theta_1 \theta_2 (x_1-x_1')(x_2-x_2') + \theta_2^2 |x_2 - x_2'|^2\bigr]}
\end{equation*} 
is a valid ARD kernel. However, if features are fully redundant, i.e. $x_1 = x_2$ and $x_1' = x_2'$, then
\begin{equation*}
       k_\theta(\x, \x') = e^{-\bigl[(\theta_1 - \theta_2)^2 |x_1-x_1'|^2 \bigr]}.
\end{equation*}
Yet, since features are equally important, we expect $\theta_1 = \theta_2$. This choice would render the kernel constant, implying that none of the features are relevant. Product ARD kernels avoid this degenerate scenario by isolating the influence of each feature in the kernel value. 

\begin{definition}[Product ARD kernel] A function $k_\theta: \sX \times \sX \to \R$ is a product ARD kernel if
\begin{equation}
    k_\theta(\x, \x')= \prod_{i=1}^d k_{\theta_i}(\sx_i, \sx_i') = \prod_{i=1}^d G[ \theta_i (\sx_i- \sx_i')],
    \label{eq:product-ard}
\end{equation}
where each $k_{\theta_i}$ is an ARD kernel of the form \eqref{eq:ard} with relevance $\theta_i \in \R_+$.
\end{definition}

We remark that limiting our attention to product ARD kernels is not restrictive. The most commonly used kernels in the literature are product ARD kernels constrained by the condition that $\theta_i = \theta_j = 1/\sigma^2$ for all $i, j \in [d]$, where $\sigma^2$ is a bandwidth parameter. As we will see, product ARD kernels are crucial to ensure \rffnet{}'s computational efficiency.

\section{Overview of \rffnet}
\label{sec:rffnet}

In this section, we introduce \rffnet{}, our large-scale and interpretable kernel method based on random Fourier features for product ARD kernels, and carefully initialized first-order optimization methods. First, we discuss two essential properties of product ARD kernels. Then, we discuss \rffnet{} objective function and how we train \rffnet{}. We conclude by introducing an algorithm for variable selection based on thresholding the relevance vector output by an \rffnet{} instance. All proofs are deferred to~\ref{appendix:proofs}.

\subsection{Random Fourier features for product ARD kernels}
\label{subsec:rff-ard}

Two key properties of ARD kernels render them particularly well-suited to a random Fourier features approximation. The first property is a simple consequence of the pairing of features and relevances in general ARD kernels. In fact, let $k_\theta$ be an ARD kernel, then
\begin{equation*}
    k_\theta(\x, \x') = G[ \theta \circ (\x - \x')] = G[ \mathbf{1}_d \circ (\theta \circ \x - \theta \circ \x')] = k_{\mathbf{1}_d} (\theta \circ \x, \theta \circ \x').
\end{equation*}
Thus, any random Fourier features map $\z: \R^d \to \R^s$ that approximates $k_{\mathbf{1}_d}$ can be equally used to approximate $k_\theta$ without compromising the statistical properties of the approximation. The next proposition formalizes this. 
\begin{proposition} 
\label{prop:kernel-approx}
Let $k_\theta: \sX \times \sX \to \R$ be an ARD kernel. Let $\z: \R^d \to \R^s$, $s \ge 1$, be the random Fourier features map for $k_{\mathbf{1}_d}$, the isotropic version of $k_\theta$. Then,
\begin{equation}
    \widehat{k}_\theta(\x, \x') = \z(\theta \circ \x)^\t \z(\theta \circ \x'),
    \label{eq:ard-kernel-rff}
\end{equation}
is an unbiased estimator of $k_\theta(\x, \x')$.
\end{proposition}

In practice, the most significant consequence of this proposition is that to approximate an ARD kernel $k_\theta$, we need to sample the spectral density of $k_{\mathbf{1}_d}$, which is independent of the relevances. The relevance vector can be introduced later as a scaling of the features. In light of the discussion in Section~\ref{subsec:ard}, this greatly reduces the computational cost of including the relevance vector in empirical risk minimization procedures, as changes in the relevance during the minimization do not require resampling the entire pool of random features. This property has been already highlighted in \citet{gregorova2018large}. 

The second property, intrinsic to product ARD kernels, stems from the form of their spectral density, characterized in the following proposition.

\begin{proposition}\label{prop:spectral-dens}  Let $k_\theta: \sX \times \sX \to \R$ be a product ARD kernel. Then, its spectral density $p_\theta$ satisfies
\begin{equation}    p_\theta(\omega_1, \dots, \omega_d) = \prod_{i=1}^d \frac{1}{|\theta_i|} p \left(\frac{\omega_i}{\theta_i}\right),
    \label{eq:product-ard-density}
\end{equation}
where $p(\cdot)$ is the spectral density of $k_{1}$, that is $k_{\theta_i}$ with $\theta_i = 1$.
\end{proposition}

This result shows that each relevance $\theta_i$ is a scale parameter of the corresponding spectral density $p_i$.  As such, the sign of each $\theta_i$ is irrelevant. When including $\theta$ in \rffnet{}'s objective function, we can treat the relevances as an unconstrained parameter \footnote{To clarify with an analogy: the standard error is a scale parameter of the normal distribution. Our assertion is analogous to stating that if $Z \sim \mathsf{N}(0, 1)$ and $\sigma > 0$, then both $\sigma Z$ and $-\sigma Z$ are distributed as $\mathsf{N}(0, \sigma^2)$.}.
In contrast, other methods restrict the values of $\theta$ to specific subsets of $\R^d_+$, such as simplices \citep{gregorova2018large, chen2017kernel} and product of intervals \citep{allen2013automatic}. Imposing these constraints requires projecting $\theta$ each time it is updated, hindering its interpretation of a vector of lengthscales of variation, which, by definition, should be unbounded.

\subsection{The objective function and training}

\rffnet{} builds on the properties above of product ARD kernels to add a layer of interpretability to kernel methods with minimal computational overhead. For a given user-specified product ARD kernel $k_\theta$ whose spectral density can be sampled, \rffnet{} solves
\begin{equation}
      \min_{\beta \in \R^s,\, \theta \in \R^d }  \frac{1}{n} \sum_{i=1}^n \ell \left( y_i,   \z(\theta \circ \x_i)^\t \beta \right) + \lambda \norm{\beta}_2^2,
      \label{eq:rffnet-problem}
\end{equation}
where $\z: \R^d \to \R^s$ is the random Fourier features map corresponding to $k_{\mathbf{1}_d}$, $\lambda > 0$ is the regularization parameter, and the loss function $\ell$ is taken as the squared loss in regression and the cross-entropy loss in classification. Notice that this is a minor modification of  \eqref{eq:approx-problem} incorporating the outcomes of Propositions~\ref{prop:kernel-approx} and~\ref{prop:spectral-dens}: the random features map is sampled only at initialization, as it does not depend on $\theta$, and $\theta$ is included in the objective as an unconstrained optimization variable. 

\rffnet{}'s objective function \eqref{eq:rffnet-problem} is highly non-convex due to the oscillatory behavior of the random features map and the parity symmetry with respect to $\theta$. Even if the objective might exhibit a favorable optimization landscape with ``natural'' input distributions, as discussed in~\ref{appendix:convexity}, the landscape with real-world data, notably in the small sample setting, is affected by these random features' oscillations. 
Nevertheless, for the aforementioned loss functions, the objective has Lipschitz continuous gradients in each block ($\beta$ and $\theta$) of coordinates. This regularity suggests we solve \eqref{eq:rffnet-problem} using carefully initialized first-order optimization methods. We give a meta-algorithm describing how \rffnet{} is trained in Algorithm~\ref{algo:rffnet}. Subsequently, we describe the default \textsc{Sample}, \textsc{Initialize}, and \textsc{Optimize} steps. Importantly, these defaults were only established after analyzing the ablation studies in \ref{appendix:ablation}. 

\begin{algorithm}[H]
\caption{Default training of \rffnet{}}\label{algo:rffnet}
\begin{algorithmic}[1]
    \Function{Fit}{training data $\mathcal{D} = \{(\x_i, y_i)\}_{i=1}^n$, number of random features $s$, regularization strength $\lambda$, max optimization iterations $T$, learning rate $\eta$, batch size $B$, validation fraction $\xi$, patience $K$} 

    \State Sample $\{b_i\}_{i=1}^s \sim \textsf{Unif}(0, 2\pi)$

    \State $\{ \omega_i \}_{i=1}^s \gets \textsc{Sample}(s)$

    \State  Use $\{(\omega_i, b_i)\}_{i=1}^s$ to construct $\z: \R^d \to \R^s$ as in~\eqref{eq:rff-entries}.

    \State $\phantom{(\beta^*}\theta_{\text{init}}\,\gets \textsc{Initialize}(\mathcal{D}) $\Comment{See~\eqref{eq:theta-init}}
    \State $(\beta^*, \theta^*) \gets \textsc{Optimize}(\mathcal{D}, \z, \theta_{\text{init}}, T, 
    \lambda, \eta, B, \xi, K)$\Comment{See Algorithm~\ref{algo:rffnet-optimize}}

    \State \textbf{return} regressor or scorer $ f^*(\x) = \z(\theta^* \circ \x)^\t \beta^*.$

    \EndFunction 
\end{algorithmic}
\end{algorithm}


In the default setting, the \textsc{Sample} step samples from the spectral density of the Gaussian isotropic kernel, i.e. $\omega_i \sim \mathsf{N}(0, I_{d})$ in step 3 of Algorithm~\ref{algo:rffnet}. With this, according to Proposition~\ref{prop:kernel-approx}, the random features map $\z$ of step 4 is such that $\widehat{k}_\theta(\x, \x')$ is an unbiased estimator of the Gaussian ARD kernel
\begin{equation*}
    k_\theta(\x, \x') = \exp\Biggl[ -\frac{1}{2} \norm{\theta \circ (\x-\x')}_2^2 \Biggr].
\end{equation*}

The relevance vector is initialized in the \textsc{Initialize} step as 
\begin{equation}
        \theta_{\text{init}}= \frac{1}{d} \Bigl( \max \{\x_i : i \in [n]\} - \min \{\x_i : i \in [n]\} \Bigr),
        \label{eq:theta-init}
    \end{equation}
where the $\max$ and $\min$ are element-wise, i.e. they act in each entry of the vectors $\x_i$. Intuitively, assuming the training data is standardized, features with the greatest range (i.e., the difference between the maximum and minimum) should vary in smaller lengthscales and thus be initialized with greater relevances. 

For the \textsc{Optimize} step, we use the Adam optimizer \citep{kingma2014adam} with mini-batches, treating $\theta$ and $\beta$ as a single block of coordinates. The $\ell_2$ penalty on $\beta$ in~\eqref{eq:rffnet-problem} is applied by composing the iterates of $\beta$ with the proximal operator of the $\ell_2$ norm
\begin{equation}
\label{eq:prox-l2}
          \prox_{\lambda \norm{\cdot}_2^2, \eta}(\beta) = \frac{\beta}{1+2 \lambda \eta} ,
\end{equation}
where $\eta$ is the learning rate. We used proximal operators to allow users to easily implement and experiment with other regularization terms on $\beta$ \textit{and} $\theta$. Finally, we used early stopping and checkpointing as training heuristics. Details of the optimization procedure are described in Algorithm~\ref{algo:rffnet-optimize}.

\begin{algorithm}[!ht]
\caption{Minimizing \rffnet{}'s objective}\label{algo:rffnet-optimize}
\begin{algorithmic}[1]
    \Function{Optimize}{$\mathcal{D}, \theta_{\text{init}}, T, 
    \lambda, \eta, B, \xi, K$}
    \State Validation split: $ \mathcal{D}' \gets \xi$ fraction of samples of $\mathcal{D}$.
    \State $\theta \gets \theta_{\text{init}}$.\Comment{See~\eqref{eq:theta-init}}.
    \For{$i \in \{1, \dots, T\}$}
        \State Sample batches of size $B$ from $\mathcal{D} \setminus \mathcal{D}' $.
        \For{each batch}
               \State $(\beta, \theta) \gets$ Adam$(\beta, \theta, \eta)$.
               \State $\beta \gets \prox_{\lambda \norm{}_2^2, \eta} (\beta) .$\Comment{See~\eqref{eq:prox-l2}}.
        \EndFor
        \State Evaluate the loss of $ f^{(i)}(\x) = \z(\theta^{(i)} \circ \x)^\t \beta^{(i)}$ on $\mathcal{D}'$.
        \State Store $(\beta^{(i)}, \theta^{(i)})$.
        \If{loss on $\mathcal{D}'$ reduced in every previous $K$ iterations}
            \State Stop training.\Comment{Early stopping.}
        \EndIf
    \EndFor
    \State $(\beta^*, \theta^*) \gets $ $(\beta^{(i)}, \theta^{(i)})$ with the smallest loss on $\mathcal{D}'$.\Comment{Checkpointing.}
    \State \textbf{return} $(\beta^*, \theta^*)$
    \EndFunction 
\end{algorithmic}
\end{algorithm}

As for the implementation, we remark that solving~\eqref{eq:rffnet-problem} is equivalent to solving a regularized loss minimization problem in the hypothesis space $\sF = \{ \x \mapsto \beta^\t \z(\theta \circ \x): \beta \in \R^s, \theta \in \R^d \}$. Since $z_i(\x) = \sqrt{2/s} \cos (\omega_i^\t \x + b_i)$, $\sF$ is the space of one hidden-layer neural networks with cosine activation functions. We use this equivalence to implement \rffnet{} as a neural network model in PyTorch and leverage PyTorch's efficient automatic differentiation engine to compute the gradients in Algorithm~\ref{algo:rffnet-optimize}.

\subsection{Thresholding for variable selection}
\label{subsec:thresholding}

The relevance vector $\theta$ output by \rffnet{} is not \textit{exactly} sparse. Even if entries corresponding to irrelevant features are shrunk during training, they are not forced to be identically zero. This does not oppose our purpose of creating an interpretable kernel method.  We expect the relevances to be powerful tools for feature engineering and model interpretation, giving insights about the relative significance of features within the given predictive task. Nevertheless, it is natural to ask if thresholding the entries of the relevance vector can yield a valid variable selection rule, capable of identifying the active features while controlling the false discovery rate.

To this end, we introduce TopK, a simple procedure to automatically include/exclude variables from a trained \rffnet{} model based on the model's scores (under a suitable scoring function) on a held-out dataset. We described TopK in Algorithm~\ref{algo:topk}. The scoring rule in Algorithm~\ref{algo:topk} can be chosen as the negative squared error loss for regression and the AUC for classification problems.

\begin{algorithm}[!ht]
\caption{Variable selection using \rffnet{}'s relevances}
\begin{algorithmic}[1]
\Function{TopK}{selection dataset $\mathcal{D}_{\text{s}}$, fitted \rffnet{} model $f^*$ with relevance vector $\theta^*$, scoring function $r: \sY \times \sY \to \R$}
\State  Order the relevances in descending order.
\For{$i \in [d]$}
    \State $\theta_{\text{thresh}} \gets \bigl[\theta^*_{(d)}, \theta^*_{(d-1)}, \dots, \theta^*_{(d-i)}, 0, \dots, 0\bigr].$\Comment{Include the $i$ greatest relevances}
    \State $\theta^* \gets \theta_{\text{thresh}}$.
    \State Score $f^*$ with the new $\theta^*$ in $\mathcal{D}_{\text{s}}$. 
    \State Store the value of the score, say $r_i$ and the vector $\theta_{\text{thresh}}$.
\EndFor
\State Get the relevance vector $\theta_{\text{thresh}}^*$ associated with the highest score.
\State $\theta_{\text{s}} \gets \theta_{\text{thresh}}^*$.
\State \textbf{return} $\theta_{\text{s}}$.
\EndFunction
\end{algorithmic}
\label{algo:topk}
\end{algorithm}

\section{Results}
\label{sec:applications}

In this section, we report the results of the evaluation of \rffnet{} in simulated and real-world regression and classification datasets. We include here only predictive metrics (mean squared error for regression and AUC for classification), leaving results about RAM usage and run time to \ref{appendix:experiments}.

\paragraph*{Benchmarks} For regression, we compared \rffnet{} to kernel ridge regression (KRR) with an isotropic Gaussian kernel, approximate kernel ridge regression with Fastfood \citep{le2013fastfood} and Nyström \citep{yang2012nystroem, williams2000using, drineas2005nystrom} feature maps for the isotropic Gaussian kernel, Gaussian Processes Regression (GPR) \citep{williams2000using} with an ARD Gaussian kernel, EigenPro regressor \citep{ma2017diving} with a Gaussian isotropic kernel, and to Sparse Random Fourier Features (SRFF) \citep{gregorova2018large} with a Gaussian kernel.
For classification, we compared \rffnet{} to logistic regression with Fastfood and Nyström feature maps for the isotropic Gaussian kernel, and to the EigenPro classifier with a Gaussian isotropic kernel. Among these benchmarks, only SRFF and GPR output feature relevances.

\paragraph*{Datasets} We compared the above benchmarks against \rffnet{} in 4 simulated regression datasets and 2 simulated classification datasets. For regression, the datasets \textsf{gse1} and \textsf{gse2} originate from \citet{gregorova2018large}, while \textsf{jse2}  and \textsf{jse3} originate from \citet{jordan2021self}. These datasets were used to evaluate the ability of kernel-based algorithms to identify active features in the simulated regression function. For classification, the \textsf{classification} and \textsf{moons} are native to \textit{scikit-learn} \citep{pedregosa2011scikit}, where they are widely used to evaluate the performance of classification algorithms. 

To complete our assessment, we evaluated \rffnet{} in 8 real-world datasets (4 regression and 4 classification datasets). These datasets comprise a range of sample sizes ($n$) and dimensionality ($d$) that allow us to pinpoint the settings where \rffnet{} is expected to perform better. A complete description of the simulated and real-world datasets and the sources to download the real-world datasets is provided in \ref{appendix:datasets}. We also provide code to preprocess these datasets and fully reproduce the results in this section. 

\paragraph*{Hyperparameters} \rffnet{} was evaluated using its default configuration with a regularization parameter of $\lambda = 10^{-4}$ motivated by the analysis in ~\ref{appendix:ablation}. Benchmarks were evaluated using their default hyperparameters. For a fair comparison, we do not optimize (e.g., via cross-validation) the hyperparameters of \rffnet{} nor any of the benchmarks. Notwithstanding, we strongly encourage users to include hyperparameter tuning methods in their data analysis pipelines. 

\subsection{Simulations}
\label{sec:simulations}

\begin{figure}[!htb]
    \centering
    \includegraphics{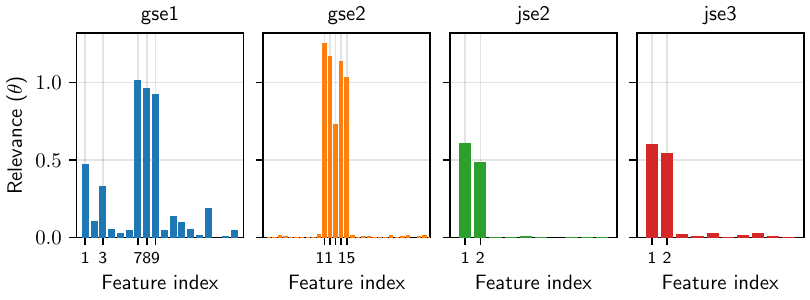}
    \caption{Relevances for a realization of the \textsf{gse1, gse2, jse2} and \textsf{jse3} simulated datasets. The labels on the x-axis indicate the active features for each dataset (see~\ref{appendix:datasets}). The relevances output by \rffnet{} peak exactly on the active features of these datasets.}
    \label{fig:relevance-simulated}
\end{figure}

First, we compared all benchmarks in the simulated datasets. We fixed the sample size as $n = 5\,000$ and generated 10 replicas of each dataset. We report the average and one standard error of the performance metrics in these replicas. In Table~\ref{tab:simulated-regression}, we notice that \rffnet{} outperforms all benchmarks by a large margin in the regression datasets. Figure~\ref{fig:relevance-simulated} helps explain this: \rffnet{} correctly identifies the active features in each dataset, removing the spurious influence of the irrelevant ones.  Table~\ref{tab:simulated-classification} shows that all methods have comparable performance for the classification datasets. This occurs because both datasets have decision boundaries that are learnable using the nonlinear feature maps used in the benchmark models. Finally, the run-time and RAM measurements in~\ref{appendix:measurements} show that fitting the relevances did not incur additional computational overhead for \rffnet{}, which outplayed GPR and SRFF.

\begin{table}[!hbt]
    \centering
    \caption{Mean squared error on the hold-out sample for the simulated regression datasets. \rffnet{} outperformed all baselines. The best performances for each metric are displayed in boldface.}
    {\footnotesize
\begin{tabular}{lllll}
\toprule
Model & \textsf{gse1} & \textsf{gse2} & \textsf{jse2} & \textsf{jse3} \\
\midrule
EigenPro & $0.100 \pm 0.004$ & $4.489 \pm 0.287$ & $\mathbf{2.652 \pm 1.759}$ & $0.037 \pm 0.005$ \\
Fastfood & $0.082 \pm 0.004$ & $5.112 \pm 0.285$ & $10.381 \pm 2.100$ & $0.055 \pm 0.011$ \\
GPR & $0.081 \pm 0.004$ & $5.134 \pm 0.269$ & $26.718 \pm 4.239$ & $0.661 \pm 0.036$ \\
Kernel Ridge & $0.091 \pm 0.004$ & $4.085 \pm 0.260$ & $\mathbf{2.018 \pm 1.585}$ & $0.032 \pm 0.005$ \\
Nyström & $0.083 \pm 0.004$ & $5.084 \pm 0.288$ & $7.031 \pm 2.211$ & $0.147 \pm 0.010$ \\
\rffnet{} & $\mathbf{0.073 \pm 0.004}$ & $\mathbf{1.865 \pm 0.222}$ & $\mathbf{1.359 \pm 1.386}$ & $\mathbf{0.012 \pm 0.002}$ \\
SRFF & $0.084 \pm 0.004$ & $5.509 \pm 0.248$ & $29.996 \pm 13.526$ & $0.227 \pm 0.052$ \\
\bottomrule
\end{tabular}}
\label{tab:simulated-regression}
\end{table}

\begin{table}[!ht]
    \centering
    \caption{Mean classification AUC on the hold-out sample for the simulated classification datasets. On average, \rffnet{} outperforms all baselines. The best performances for each model are displayed in boldface. Since all models considered fit non-linear decision boundaries, they perfectly fit the simple \textsf{moons} dataset.}
    {\footnotesize 
\begin{tabular}{lll}
\toprule
Model & \textsf{classification} & \textsf{moons} \\
\midrule
EigenPro & $0.959 \pm 0.023$ & $1.00 \pm 0.00$ \\
Fastfood & $0.960 \pm 0.018$ & $1.00 \pm 0.00$ \\
Nyström & $0.964 \pm 0.020$ & $1.00 \pm 0.00$ \\
\rffnet{} & $\mathbf{0.98 \pm 0.016}$ & $\mathbf{1.00 \pm 0.00}$ \\
\bottomrule
\end{tabular}}
\label{tab:simulated-classification}
\end{table}


\subsection{Variable selection} 
\label{sec:variable-selection}
\begin{figure}[!ht]
    \centering
    \includegraphics{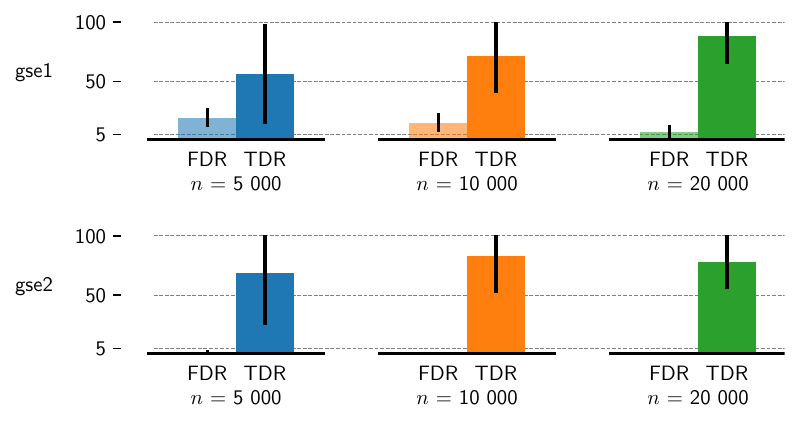}
    \caption{The empirical False/True Discovery Rates (FDR/TDRs) for the simulated datasets, \textsf{gse1} and \textsf{gse2}. Increasing the sample size controls the FDR and improves the identification of active features.}
    \label{fig:selection}
\end{figure}

We test whether TopK (see Algorithm~\ref{algo:topk} yields a variable selection rule with a high True Discovery Rate (TDR) that controls the False Discovery Rate (FDR). We used the negative of the squared error loss as the scoring function. In Figure~\ref{fig:selection}, we tested TopK for increasing sample sizes in the \textsf{gse1} and \textsf{gse2} datasets. We report the average of each metric, along with the standard error, over 10 replicas of each dataset. Observe that increasing the sample size improves the identification of the relevant features, as the mean TDR increases while the variance decreases. In addition, the FDR rate is kept low in all situations and monotonically reduces as the sample size increases.

\subsection{Real-world datasets}

\begin{figure}[!htb]
\begin{subfigure}[b]{\textwidth}
    \centering
    \includegraphics[scale=0.9]{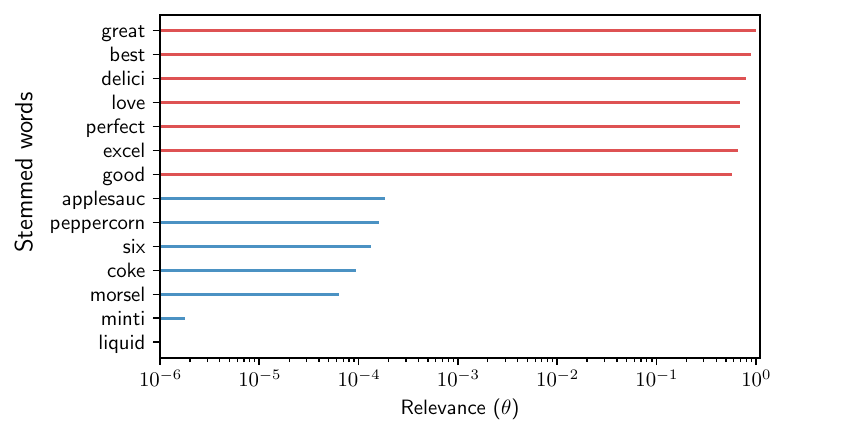}
    \caption{Relevances for the top 7 most relevant (in red) and 7 least relevant (in blue) features for the \textsf{amazon} dataset. Notice how the relevant features are, indeed, more informative to evaluate the sentiment of a food review.}
\end{subfigure}
\begin{subfigure}[b]{\textwidth}
    \centering
    \includegraphics[scale=0.9]{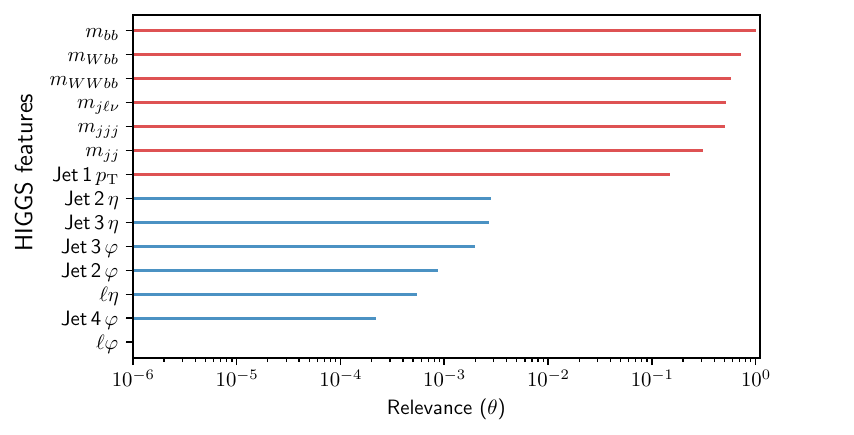}    
    \caption{Relevance for the top 7 most relevant (in red) and 7 least relevant (in blue) features for the \textsf{higgs} dataset. The most relevant features are exactly the high-level features in \citet{baldi2014searching} that better distinguish Higgs-related events from background ones.}
\end{subfigure}
    \caption{Relevance patterns for the \textsf{amazon} (top) and \textsf{higgs} (bottom) datasets. \rffnet{} associates greater relevance to features that are intuitively (for the \textsf{amazon} dataset) and scientifically relevant (for the \textsf{higgs} dataset).}
    \label{fig:relevance-real}
\end{figure}

Finally, we report the results of applying \rffnet{} to real data. We compared all benchmarks via 10-fold cross-validation and reported the average and standard error of the prediction metric in the held-out sample. In Table~\ref{tab:real-regression}, we show the results for the regression datasets. We observe that \rffnet{} substantially outperforms SRFF, which did not converge for 3 out of 4 datasets used, and GPR, which incurred large errors for most tasks. \rffnet{} performed similarly to all benchmarks in the \textsf{abalone} dataset and outperformed all methods in the \textsf{compact} dataset. In the \textsf{powerplant} dataset, we believe that the low number of features discouraged the convergence of \rffnet{} to a lower error solution. The results in~\ref{appendix:measurements} show that \rffnet{} maintained both a small run-time and memory usage due to the random Fourier features approximation. Table~\ref{tab:real-classification} summarizes the result for the classification datasets. \rffnet{} significantly outperformed the benchmarks in the \textsf{amazon} and \textsf{higgs} datasets. This is the result of correctly identifying relevant features for the predictive task, as depicted in Figure~\ref{fig:relevance-real}. In the \textsf{amazon} dataset, where the objective is to predict the sentiment (positive or negative) of a food review, the most relevant features intuitively have strong correlations with positive sentiments. In the \textsf{higgs} dataset, where the goal is to distinguish between physical processes leading to the production of Higgs bosons to background processes, the most relevant features are the high-level covariates created in \citet{baldi2014searching} to have greater distinguishing power between these processes. Importantly, as discussed in the supplementary results in \ref{appendix:measurements}, \rffnet{} superior's predictive performance in these datasets do not come at prohibitive run-time and RAM usage.  

\begin{table}[!htb]
    \centering
    \caption{Mean squared error on the hold-out sample for the real-world regression datasets. The best performances for each model are displayed in boldface. OOM: out-of-memory (see \ref{appendix:experiments}). FC: failed to converge.}
    {\footnotesize 
\begin{tabular}{lllll}
\toprule
Model & \textsf{abalone} & \textsf{compact} & \textsf{powerplant} & \textsf{yearprediction} \\
\midrule
EigenPro & $\mathbf{4.449 \pm 0.402}$ & $42.546 \pm 5.701$ & $17.664 \pm 0.888$ & OOM \\
Fastfood & $\mathbf{4.507 \pm 0.347}$ & $41.995 \pm 4.924$ & $17.031 \pm 1.06$ & $88.156 \pm 1.043$ \\
GPR & $\mathbf{4.904 \pm 0.391}$ & $540.443 \pm 37.107$ & $41.79 \pm 4.81$ & OOM \\
Kernel Ridge & $\mathbf{4.447 \pm 0.406}$ & $35.3 \pm 5.669$ & $\mathbf{15.555 \pm 0.959}$ & OOM \\
Nyström & $\mathbf{4.408 \pm 0.396}$ & $38.43 \pm 10.196$ & $\mathbf{15.092 \pm 0.939}$ & $\mathbf{81.421 \pm 1.106}$ \\
\rffnet{} & $\mathbf{4.851 \pm 0.417}$ & $\mathbf{9.765 \pm 1.254}$ & $29.762 \pm 1.636$ & $95.367 \pm 1.385$ \\
SRFF & $5.267 \pm 0.419$ & FC & FC & FC \\

\bottomrule
\end{tabular}}
\label{tab:real-regression}
\end{table}

\begin{table}[!ht]
    \centering
\caption{Mean classification AUC on the hold-out sample for the real-world classification datasets. \rffnet{} performs better in large-scale datasets. The best performances for each model are displayed in boldface.}
    {\footnotesize 
\begin{tabular}{lllll}
\toprule
Model & \textsf{a9a} & \textsf{amazon} & \textsf{higgs} & \textsf{w8a} \\
\midrule
EigenPro & $0.867 \pm 0.006$ & OOM & OOM & $0.956 \pm 0.010$ \\
Fastfood & $\mathbf{0.899 \pm 0.004}$ & $0.867 \pm 0.005$ & $0.701 \pm 0.004$ & $0.947 \pm 0.014$ \\
Nyström & $0.894 \pm 0.004$ & $0.825 \pm 0.007$ & $0.723 \pm 0.002$ & $\mathbf{0.967 \pm 0.008}$ \\
\rffnet{} & $0.890 \pm 0.004$ & $\mathbf{0.922 \pm 0.002}$ & $\mathbf{0.774 \pm 0.002}$ & $0.949 \pm 0.008$ \\
\bottomrule
\end{tabular}}
\label{tab:real-classification}
\end{table}

\section{Final remarks}
\label{sec:final}

In this paper, we introduced \rffnet{}, a new kernel method that yields interpretable prediction rules for large-scale datasets. \rffnet{} reliance on carefully designed random Fourier features and stochastic first-order methods grants low run-time and memory usage even for datasets exceeding millions of samples. Additionally, \rffnet{}'s defaults are established through comprehensive ablation studies, providing users with a sensible starting point for applying the method to their data.

Moreover, automatic relevance determination kernels bridge the gap between kernel methods' nonparametric, high-dimensional nature and the practitioner's necessity for interpretable solutions based on the original features. The kernel relevances give users a reliable metric of feature importance, revealing the presence of irrelevant covariates and underscoring that active features may have distinct length scales of variation. 

We validated our approach in a series of experiments, wherein \rffnet{} exhibited a performance on par with other kernel methods, notably in the large sample scenario with many inactive features. In the simulated datasets, \rffnet{} showed promising predictive performance and precise ability to identify the relevant features. For the real-world datasets, we observed that \rffnet{} consistently outperforms or matches the performance of many kernel-based learning algorithms. 

Many open venues for future work remain. On the theory side, does the inclusion of the relevances in the empirical risk minimization procedure drastically increase the problem's sample complexity? Furthermore, can we use the relevances during training to reduce the number of parameter updates, akin to active set methods in sparse linear regression? If possible, this could improve \rffnet{} scalability even further. On the applications side, we want to leverage \rffnet{} layered structure to incorporate it in models like neural networks, aiming for interpretable prediction on structured inputs such as images and tensors.
 
\section*{Acknowledgements}
We thank Carlos Pagani Zanini, Eduardo Fonseca Mendes, Júlio Michael Stern, and Roberto Imbuzeiro for helpful suggestions during the early stages of this work. MPO thanks Flávia Castro Motta for attentively reviewing the paper. MPO was supported through grant 2021/02178-8, São Paulo Research Foundation (FAPESP). RI was supported through grants 309607/2020-5, 422705/2021-7, 305065/2023-8 Brazilian National Counsel of Technological and Scientific Development (CNPq), and grants 2019/11321-9, 2023/07068-1, São Paulo Research Foundation (FAPESP).

\appendix

\section{Proofs}
\label{appendix:proofs}

\begin{lemma}[Relation between spectral densities of $k_\theta$ and $k_{\mathbf{1}_d}$] Let $p_\theta$ be the spectral density of the PSD shift-invariant kernel $k_\theta: \sX \times \sX \to \R$ and $p$ be the spectral density of $k_\theta$ with $\theta = \mathbf{1}_d$. Then,
\begin{align}
    p_\theta(\omega) =\frac{1}{| \theta_1 \cdots \theta_d| } p\left( \omega \circ \frac{1}{\theta} \right),\label{eq:density}
\end{align}
where $\omega \circ 1/\theta = (\omega_1/\theta_1, \dots, \omega_d/\theta_d)$.
\end{lemma}
\begin{proof} The density of the spectral measure of $k_\theta$ is, by Bochner's Theorem,
\begin{align}
    p_\theta(\omega) &= \frac{1}{(2\pi)^d} \int e^{-i \omega^\t (\x-\x')} k_\theta(\x,\x') d \delta \nonumber \\ 
    &= \frac{1}{(2\pi)^d} \int e^{-i \omega^\t \delta} G(\theta \circ \delta)  d \delta, \nonumber 
\end{align}
where $\delta = \x-\x'$. Define the new variable $\delta' = \theta \circ \delta = (\theta_1 \delta_1, \dots, \theta_d \delta_d)$, then, by the multivariate change of variables theorem, we get
\begin{align}
    p_\theta(\omega) &= \frac{1}{(2\pi)^d} \int e^{ - i \left(\omega \circ \frac{1}{\theta} \right)^\t \delta' } G(\delta') \frac{1}{| \theta_1 \cdots \theta_d| } d\delta' \nonumber \\ 
    &= \frac{1}{| \theta_1 \cdots \theta_d| }  \frac{1}{(2\pi)^d}  \int e^{ - i \left(\omega \circ \frac{1}{\theta} \right)^\t \delta' } G(\delta') d\delta' \nonumber \\ 
    &=\frac{1}{| \theta_1 \cdots \theta_d| } p\left( \omega \circ \frac{1}{\theta} \right),
\end{align}
with $p(\cdot)$ the spectral measure of the kernel with $\theta = \mathbf{1}_p$.
    
\end{proof}

\begin{proof}[Proof of Proposition 1] 

By \eqref{eq:density}, if we sample $\omega \sim p(\cdot)$, then $\theta \circ \omega \sim p_\theta(\cdot)$. Now, by Bochner's Theorem, with $\omega' \sim p_\theta(\cdot)$,
\begin{align*}
    k_\theta(\x, \x') &= \E_{\omega'}[ \cos \omega'^\t (\x- \x')]  \\ 
    &= \E_{\omega}[ \cos (\theta \circ \omega)^\t (\x - \x')]  \\ 
    &= \E_{\omega}[ \cos \omega^\t \left( \theta \circ (\x- \x') \right) ],
\end{align*}
where $\omega \sim p(\cdot)$. Let $\z: \R^d \to \R^s$ be the RFF for $k_\theta$ with $\theta = \mathbf{1}_d$, then
\begin{equation*}
  k_\theta(\x, \x') = \E_{\omega, b}[ \z(\theta \circ \x)^\t \z(\theta \circ \x')],
\end{equation*}
which shows that
\begin{equation*}
    \widehat{k}_\theta(\x, \x') =  \z(\theta \circ \x)^\t \z(\theta \circ \x')
\end{equation*}
is an unbiased estimator of $k_\theta(\x, \x')$.
\end{proof}

\begin{proof}[Proof of Proposition 2] Since $k_\theta$ is a product ARD kernel
\begin{equation*}
    k_\theta(\x, \x') = \prod_{i=1}^d k_{\theta_i}(x_i, x'_i) =  \prod_{i=1}^d G[\theta_i(x_i - x'_i)],
\end{equation*}
the spectral density of $p_\theta$ reads
\begin{align*}
   p_\theta(\omega) &= \frac{1}{(2\pi)^d} \int e^{-i \omega^\t (\x-\x')} k_\theta(\x,\x') d \delta \\ 
   &= \prod_{i=1}^d  \frac{1}{2\pi |\theta_i| } \int e^{-i \frac{\omega_i}{\theta_i} \delta'_i} G(\delta'_i) d\delta'_i,
\end{align*}
where $\delta'_i = \theta_i(x_i - x'_i)$. As each term in the product is the spectral density of the one-dimensional ARD kernels $k_{\theta_i}$, we conclude that
\begin{equation*}
       p_\theta(\omega) = \prod_{i=1}^d \frac{1}{|\theta_i|} p\left(\frac{\omega_i}{\theta_i} \right),
\end{equation*}
where $p$ is the spectral density of any $k_{\theta_i}$ with $\theta_i = 1$.
\end{proof}

\section{Experimental details}
\label{appendix:experiments}

\subsection{Code availability} 

Code to reproduce the experiments is available at 
\begin{center}
    \url{https://github.com/mpotto/rffnet}
\end{center}

\subsection{Computing infrastructure}

Experiments were run with an Intel Core i7-8700 CPU with 3.20GHz, 6 cores, 12 threads, and 54 GB of RAM. GPUs were not used to facilitate the reproduction of the results. We use the OOM (out-of-memory) to indicate that a particular experiment exceeded the memory budget.

\section{Datasets}
\label{appendix:datasets}

Simulated datasets were created as follows:
\begin{itemize}

    \item \textsf{gse1} \citep{gregorova2018large}: features are sampled as $X \sim \mathsf{N}(0, I_d)$, with $d = 18$.  The noise variables are sampled as $\epsilon \sim \mathsf{N}(0, \sigma=0.1)$. The response is:
    \begin{equation*}
        y = \sin[(x_1+x_3)^2]\sin(x_7 \, x_8 \, x_9) + \varepsilon.
    \end{equation*}   

    \item \textsf{gse2} \citep{gregorova2018large}: features are sampled as $X \sim \mathsf{N}(0, I_d)$, with $d = 100$.  The noise variables are sampled as $\epsilon \sim \mathsf{N}(0, \sigma=0.1)$. The response is:
    \begin{equation*}
      y = \log[(x_{11}+x_{12}+x_{13}+x_{14}+x_{15})^2] + \varepsilon.
    \end{equation*}   

    \item \textsf{jse2} \citep{jordan2021self}: features are sampled as $X \sim \mathsf{N}(0, \Sigma)$, where $\Sigma$ is the $d \times d$ covariance matrix with $d=10$ and  $\Sigma_{ij} = 0.5^{|i-j|}$. The noise variables are sampled as $\epsilon \sim \mathsf{N}(0, \sigma=0.1)$. The response is
    \begin{equation*}
       y = x_1^3 + x_2^3 + \varepsilon.
    \end{equation*}   

    \item \textsf{jse3} \citep{jordan2021self}: features are sampled as $X \sim \mathsf{N}(0, I_d)$, with $d = 10$. The noise variables are sampled as $\epsilon \sim \mathsf{N}(0, \sigma=0.1)$. The response is
    \begin{equation*}
        y = x_1 x_2 + \varepsilon.
    \end{equation*}

    \item \textsf{classification}: this is a default binary classification synthetic dataset in \textit{scikit-learn} \citep{pedregosa2011scikit}. 

    \item \textsf{moons}: this is a default binary classification synthetic dataset in \textit{scikit-learn} \citep{pedregosa2011scikit}. 
    
\end{itemize}

\begin{table}[!htb]
    \centering
    \caption{Description of real-world datasets: sample size of train and test splits, and number of features.}
    {\footnotesize
    \begin{tabular}{l c c c } \toprule
    Dataset & Train ($n$) & Test ($n_{\text{test}}$) & Features ($d$) \\ \midrule
    \textsf{abalone} & 3\,759 & 419 & 8 \\
    \textsf{compact} & 7\,372 & 820 & 21 \\
    \textsf{powerplant} & 43\,056 & 4\,784 & 4 \\
    \textsf{yearprediction} & 463\,810 & 51\,535 & 90  \\
    \textsf{a9a} & 43\,957 & 4\,885 & 123 \\
    \textsf{w8a} & 58\,230 & 6\,470 & 300  \\
    \textsf{amazon} & 327\,753, & 36\,418 & 3000  \\
    \textsf{higgs} & 9\,900\,000 & 1\,100\,000 & 28 \\ \bottomrule 
    \end{tabular}}
    \label{tab:realdatasets-characteristics}
\end{table}

Table~\ref{tab:realdatasets-characteristics} gives a basic description of the real-world datasets, which were retrieved from the following sources:
\begin{itemize}
    \item \textsf{abalone}: \\[5pt] 
    {\small\url{https://www.csie.ntu.edu.tw/~cjlin/libsvmtools/datasets/regression/}} 

    \item \textsf{compact}: \\[5pt]
    {\small\url{https://www.dcc.fc.up.pt/~ltorgo/Regression/compact.tar.gz}} 

    \item \textsf{powerplant}: \\[5pt]
    {\small\url{https://archive.ics.uci.edu/dataset/294/combined+cycle+power+plantp}}

    \item \textsf{yearprediction}:  \\[5pt]
    {\small\url{https://www.csie.ntu.edu.tw/~cjlin/libsvmtools/datasets/regression/}}

    \item \textsf{a9a}: \\[5pt]
    {\small\url{https://www.csie.ntu.edu.tw/~cjlin/libsvmtools/datasets/binary}}
    
    \item \textsf{amazon}: \\[5pt]
    {\small\url{https://www.kaggle.com/datasets/snap/amazon-fine-food-reviews}}
    
    \item \textsf{higgs}: \\[5pt]
    {\small\url{https://archive.ics.uci.edu/dataset/280/higgs}}

    \item \textsf{w8a}: \\[5pt]
    {\small\url{https://www.csie.ntu.edu.tw/~cjlin/libsvmtools/datasets/binary}}

\end{itemize}

Scripts to download and preprocess these datasets are available in the \texttt{experiments/data} folder. 

\section{Ablation studies}
\label{appendix:ablation}

This section evaluates the impact of fixing the sampling, initialization, and optimization methods in Algorithm~\ref{algo:rffnet}. In addition, we seek to understand the influence of regularization on the model. We use the evaluation in this section to establish sensible defaults for \rffnet{}. The performance of \rffnet{} with these default settings is presented in Section~\ref{sec:applications}. Simulated datasets used in this section are described in \ref{appendix:datasets}. Importantly, all ablation studies were performed only in the regression setting using the squared error loss. 

\subsection{Sampling}
\label{subsec:sampling}

\begin{figure}[!ht]
    \centering
    \includegraphics{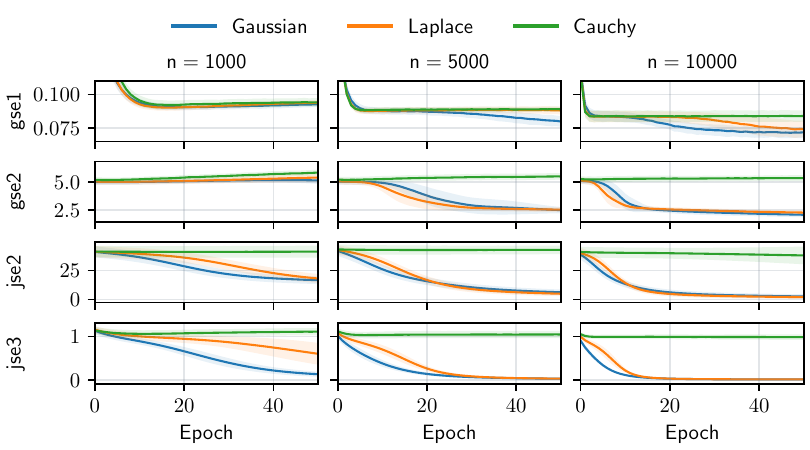}
    \caption{Validation squared error loss as a function of the training epoch for the sampling strategies. Using the Gaussian sampling scheme led to faster convergence in all cases.}
    \label{fig:ablation-samplers-reg}
\end{figure}

We tested three spectral density samplers: Gaussian, Laplace, and Cauchy. Sampling from each of these corresponds to producing an approximation of the Gaussian, Cauchy, and Laplace kernels, respectively, as outlined in Table~\ref{tab:sampling}. In this experiment, we fixed the initialization scheme as the Constant strategy described in ~\ref{subsec:initialization} and the PALM-Adam algorithm described in \ref{subsec:optimization}. The experiment is repeated 20 times. We report the average (solid lines) and 1 standard error bounds (shaded region). In Figure~\ref{fig:ablation-samplers-reg}, we see that sampling from the Gaussian and Laplace distributions often led to the same final validation loss. However, the convergence with the Gaussian sampler is usually faster than the Laplace density. For this reason, we choose the Gaussian spectral density as the default for \rffnet{}. 

\begin{table}[!hbt]
    \centering
    \caption{Usual shift-invariant PSD kernels and the densities of their corresponding spectral measures.}
    \def\arraystretch{2}
    {\footnotesize
    \begin{tabular}{l l l} \toprule 
    \textsc{Sample}  &  $k_{\mathbf{1}_d}(\x, \x')$ & $p(\omega)$  \\  \midrule 
    Gaussian & $\exp\left[-\frac{1}{2} \norm{\x-\x'}_2^2\right]$ & $(2 \pi)^{-\frac{d}{2}} \exp\left[-\frac{\norm{\omega}_2^2}{2}\right]$\\
    Laplace & $\exp\left[- \norm{\x-\x'}_1\right]$ & $\prod_{i=1}^d \frac{1}{\pi(1+\omega_i^2)}$ \\
    Cauchy &  $\prod_{i=1}^d \frac{2}{1+(x_i-y_i)^2}$ & $\exp\left[-\norm{\omega}_1\right]$ \\     \bottomrule     
    \end{tabular}}
    \label{tab:sampling}
\end{table}

\subsection{Initialization}
\label{subsec:initialization}

As the source of the objective function's non-convexity is the dependence on $\theta$, we focus on initialization strategies only for this parameter. We tested three initialization methods that users can choose from: Constant, Restarter, and Regressor. Given the training sample $\{(\x_i, y_i)\}_{i=1}^n$, these schemes work as follows:
\begin{enumerate}
    \item Constant:  the relevances are initialized as
    \begin{equation}
    \label{eq:thetastar}
        \theta_{\text{init}} = \theta_{\star} \coloneqq  \frac{1}{d} \Bigl( \max \{\x_i : i \in [n]\} - \min \{\x_i : i \in [n]\} \Bigr),
    \end{equation}
    where the $\max$ and $\min$ are feature-wise, i.e. they act in each entry of the vectors $\x_i$ (as mentioned in Section~\ref{sec:applications}).

    \item Regressor: we use $\{(\x_i, y_i)\}_{i=1}^n$ to train a Lasso \citep{tibshirani1996regression} regressor, whose output is taken as having the form $f^*(\x) = \beta^\t \x$. Then, we initialize $\theta$ with the trained model weights $\theta_{\text{init}} = \beta$.  

    \item Restarter: the relevances are first sampled as $\theta_{\text{sample}} \sim N(\theta_{\star}, \frac{1}{d} I_d)$, where $\theta_{\star}$ was defined in \ref{eq:thetastar}. Then, the model is optimized for $10$ iterations, and the final training loss is recorded. This process is restarted $10$ times from $\theta_{\text{sample}}$. $\theta_{\text{init}}$ is assigned to the relevance vector that led to the smallest training loss among the 10 runs.
\end{enumerate}

\begin{figure}[!ht]
    \centering
    \includegraphics{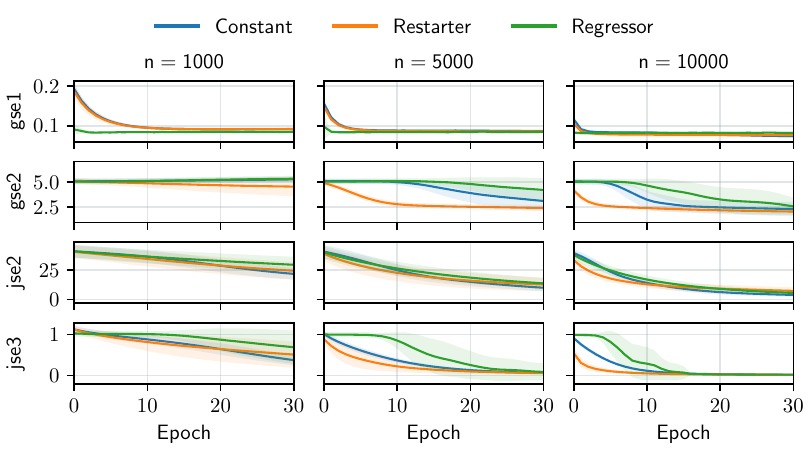}
    \caption{Validation squared error loss as a function of the training epoch for the initialization strategies. The Constant method is reliable and faster, avoiding multiple model initializations (Restarter) or training auxiliary models (Regressor).}
    \label{fig:ablation-initialization-reg}
\end{figure}

For this experiment, we fix the optimization method as the PALM-Adam algorithm described in \ref{subsec:optimization} and the sampling method as the Gaussian one described in \ref{subsec:sampling}. The experiment is repeated 20 times. We report the average (solid lines) and 1 standard error bounds (shaded region). In Figure~\ref{fig:ablation-initialization-reg}, we plot the validation loss during training for the simulated regression datasets described in \ref{appendix:datasets}. While the Restarter strategy usually gives superior initialization results, characterized by faster error decay in the early stages of training, we opt for the Constant strategy for two primary reasons. Firstly, it is much faster, as it does not require multiple training runs from distinct initializations. Secondly, the number of epochs needed for this strategy to reach the loss plateau first reached by the Restarter one is small. Finally, the Regressor strategy takes much longer to reach the same loss plateau and requires training the Lasso model at initialization, which can introduce additional computational overhead. 

\subsection{Optimization}
\label{subsec:optimization}

\begin{figure}[!ht]
    \centering
    \includegraphics{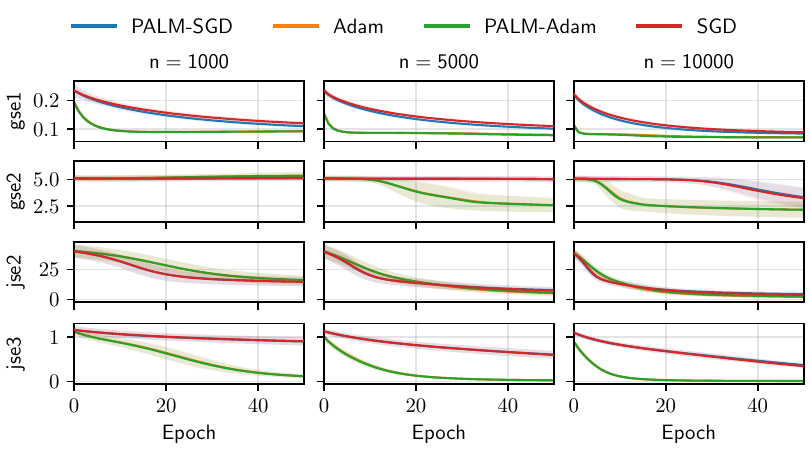}
    \caption{Validation squared error loss as a function of the training epoch for optimization algorithms. The Adam and Adam and PALM-Adam algorithms exhibit comparable performance, much superior to the SGD and PALM-SGD counterparts. Since Adam is simpler than PALM-Adam, we choose Adam as the default optimization algorithm for \rffnet{}.}
    \label{fig:ablation-solvers-reg}
\end{figure}

 We tested four first-order stochastic gradient-based optimization algorithms: vanilla Stochastic Gradient Descent (SGD) and the Adam algorithm \cite{kingma2014adam}, each with two blocks (alternating minimization) or a single block of coordinates. As described in the main text, the $\ell_2$ penalty on $\beta$ in \eqref{eq:rffnet-problem} is applied in either case by composing the iterates with the proximal operator of the $\ell_2$ norm. Because of this, the two block optimization methods are a version of the Proximal Alternating Linearized Minimization (PALM) \citep{bolte2014proximal}, while the single block methods are Proximal Stochastic Gradient Descent methods. In this section's experiments, we did not use early stopping as a training heuristic. We used the sampling method as Gaussian and initialized $\theta$ via the Constant strategy. As before, the experiments are repeated 10 times. We report the average (solid line) and 1 standard error bounds (shaded regions). We show in Figures~\ref{fig:ablation-solvers-reg} the validation loss history during training. Notice that SGD and PALM-SGD exhibited similar performance, as did Adam and PALM-Adam. However, the Adam and PALM-Adam category shows superior performance, converging much faster to a solution in the most challenging datasets (\textsf{gse1} and \textsf{gse2}). Since Adam is simpler than PALM-Adam, we opt for using Adam with a single block of coordinates.

\section{Effect of regularization strength}

\begin{figure}[!ht]
    \centering
    \includegraphics{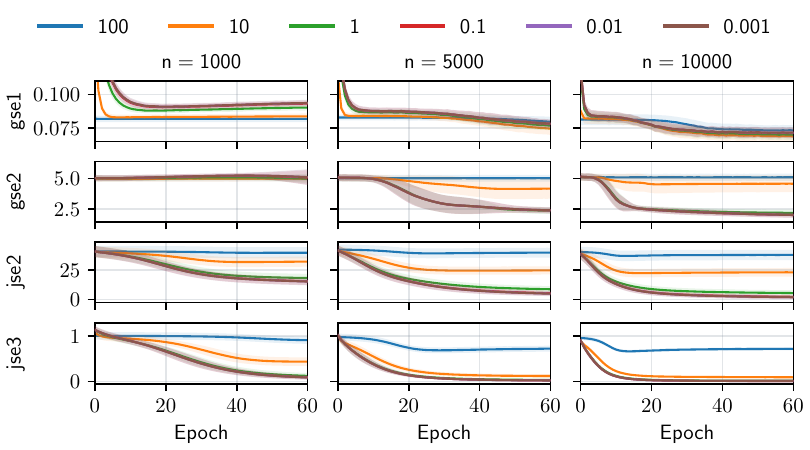}
    \caption{Validation squared error loss as a function of the training epoch for different values of the regularization parameter $\lambda$. Higher values of $\lambda$ prevent the convergence of \rffnet{} to a solution with a low validation error.}
    \label{fig:alpha-regression}
\end{figure}

\begin{figure}[!ht]
    \centering
    \includegraphics{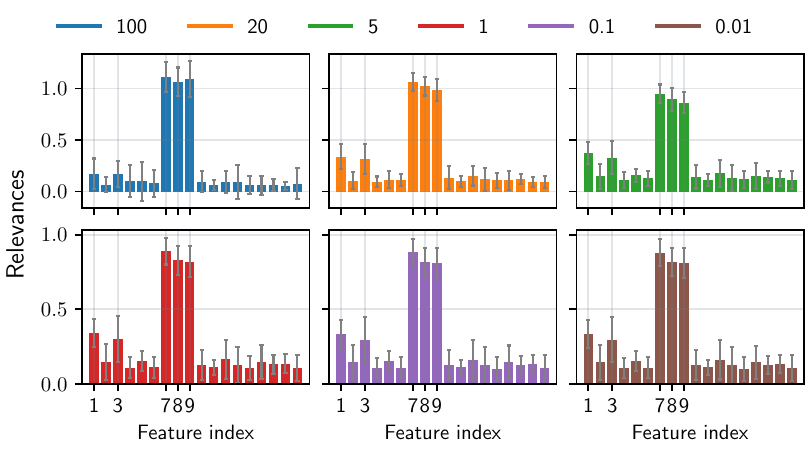}
    \caption{Relevance pattern for \textsf{gse1} dataset with $n = 5\,000$ samples and varying $\lambda$. The highest value of $\lambda$ precludes the identification of the first and third features. Reducing $\lambda$ increases the variance of the $\theta$ estimates but improves the distinction between relevant and irrelevant features.}
    \label{fig:alpha-gse1-5000}
\end{figure}

\begin{figure}[!ht]
    \centering
    \includegraphics{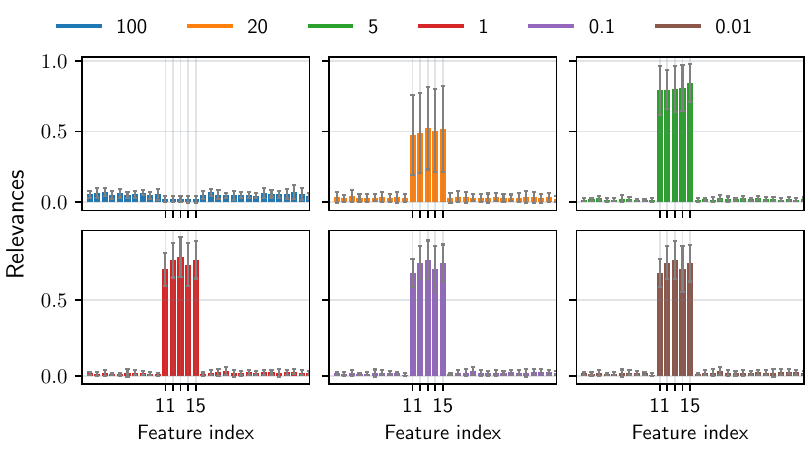}
    \caption{Relevance pattern for \textsf{gse2} dataset with $n = 5\,000$ samples and varying $\lambda$. The highest value of $\lambda$ precludes the identification of all relevant features. Reducing $\lambda$ improves the identification of features altogether.}
    \label{fig:alpha-gse2-5000}
\end{figure}

Using the default settings of \rffnet{}, we evaluated the effect of the regularization parameter $\lambda$ on the identification of the relevant features. As there may be some interaction between $\lambda$ and $\theta$ (both increasing $\lambda$ and shrinking $\theta$ increase the smoothness), we wanted to determine if there existed sensible default values for $\lambda$ that do not impose too much smoothness on the \rffnet{} predictor, unintentionally preventing the identification of the relevant features. In Figure~\ref{fig:alpha-regression}, we plot the squared error loss in the validation set during training. Notice that setting $\lambda$ too high (e.g., $\lambda \in \{10, 100\}$) precludes the convergence to a low-error solution in most cases. However, all solutions with $\lambda < 1$ attain the same validation loss plateau. What is the impact of choosing $\lambda$ on the relevance pattern identified by \rffnet{}? We analyzed this question for the \textsf{gse1} and \textsf{gse2} datasets in Figures~\ref{fig:alpha-gse1-5000} and \ref{fig:alpha-gse2-5000}. We fix the sample size as $n = 5\,000$. In both cases, we see that fixing a high $\lambda$ value prevents the identification of the first and third features in \textsf{gse1} and all the features in \textsf{gse2}. In contrast, setting smaller values of $\lambda$ may increase the variance of the relevance estimates, but does not prevent the identification of relevant features and guarantee convergence to a low validation error solution, as shown in Figure~\ref{fig:alpha-regression}. For these reasons, we have established the default regularization parameter as $\lambda = 10^{-4}$.

\section{Additional experiment results}\label{appendix:measurements}

\subsection{Run-time measurements}

\begin{table}[H]
    \centering
    \caption{Run-time (in seconds) for the simulated regression datasets.}
    {\footnotesize
\begin{tabular}{lllll}
\toprule
Model & \textsf{gse1} & \textsf{gse2} & \textsf{jse2} & \textsf{jse3} \\
\midrule
EigenPro & $3.568 \pm 0.076$ & $3.485 \pm 0.052$ & $3.85 \pm 0.031$ & $3.891 \pm 0.065$ \\
Fastfood & $0.015 \pm 0.001$ & $0.019 \pm 0.001$ & $0.014 \pm 0.001$ & $0.014 \pm 0.001$ \\
GPR & $108.689 \pm 0.185$ & $132.261 \pm 0.143$ & $106.098 \pm 0.083$ & $105.758 \pm 0.094$ \\
Kernel Ridge & $1.73 \pm 0.002$ & $1.75 \pm 0.003$ & $1.743 \pm 0.001$ & $1.747 \pm 0.004$ \\
Nyström & $0.009 \pm 0.00$ & $0.008 \pm 0.001$ & $0.009 \pm 0.00$ & $0.01 \pm 0.001$ \\
\texttt{RFFNet} & $1.838 \pm 0.62$ & $4.028 \pm 0.008$ & $3.761 \pm 0.007$ & $2.741 \pm 0.526$ \\
SRFF & $5.264 \pm 2.153$ & $34.42 \pm 9.395$ & $8.98 \pm 10.766$ & $12.714 \pm 8.075$ \\
\bottomrule
\end{tabular}}
\end{table}

\begin{table}[H]
    \centering
    \caption{Run-time (in seconds) for the simulated classification datasets.}
    {\footnotesize 
\begin{tabular}{lllll}
\toprule
Model & \textsf{classification} & \textsf{moons} \\
\midrule
EigenPro & $3.789 \pm 0.112$ & $3.625 \pm 0.108$ \\
Fastfood & $0.249 \pm 0.038$ & $0.273 \pm 0.012$ \\
Nyström & $0.189 \pm 0.021$ & $0.091 \pm 0.015$ \\
\texttt{RFFNet} & $2.805 \pm 0.979$ & $3.904 \pm 0.004$ \\
\bottomrule
\end{tabular}}
\end{table}

\begin{table}[H]
    \centering
    \caption{Run-time (in seconds) for the real-world regression datasets.  OOM: out-of-memory.}
{\scriptsize 
\begin{tabular}{lllll}
\toprule
Model & \textsf{abalone} & \textsf{compact} & \textsf{powerplant} & \textsf{yearprediction} \\
\midrule
EigenPro & $2.036 \pm 0.018$ & $3.218 \pm 0.075$ & $26.016 \pm 0.056$ & OOM \\
Fastfood & $0.006 \pm 0.001$ & $0.012 \pm 0.001$ & $0.062 \pm 0.001$ & $3.431 \pm 0.013$ \\
GPR & $17.864 \pm 0.123$ & $78.957 \pm 0.367$ & $1878.938 \pm 0.454$ & OOM \\
Kernel Ridge & $0.183 \pm 0.007$ & $0.865 \pm 0.009$ & $65.332 \pm 0.086$ & OOM \\
Nyström & $0.004 \pm 0.00$ & $0.008 \pm 0.00$ & $0.035 \pm 0.00$ & $1.806 \pm 0.003$ \\
\texttt{RFFNet} & $0.607 \pm 0.125$ & $1.707 \pm 0.571$ & $12.563 \pm 1.983$ & $84.319 \pm 15.361$ \\
SRFF & $15.343 \pm 5.339$ & $2.898 \pm 1.456$ & $71.513 \pm 72.241$ & $265.975 \pm 140.602$ \\ \bottomrule
\end{tabular}}
\end{table}

\begin{table}[H]
    \centering
    \caption{Run-time (in seconds) for the real-world classification datasets. OOM: out-of-memory.}
    {\footnotesize 
\begin{tabular}{lllll}
\toprule
Model & \textsf{a9a} & \textsf{amazon} & \textsf{higgs} & \textsf{w8a} \\
\midrule
EigenPro & $31.886 \pm 0.065$ & OOM & OOM & $89.453 \pm 0.212$ \\
Fastfood & $1.659 \pm 0.144$ & $116.829 \pm 21.278$ & $1495.365 \pm 10.466$ & $2.988 \pm 0.032$ \\
Nyström & $1.013 \pm 0.019$ & $13.208 \pm 0.676$ & $378.041 \pm 122.552$ & $1.345 \pm 0.033$ \\
\texttt{RFFNet} & $3.226 \pm 1.118$ & $109.041 \pm 40.542$ & $800.508 \pm 310.327$ & $4.023 \pm 1.625$ \\
\bottomrule
\end{tabular}}
\end{table}

\subsection{Memory measurements}

\begin{table}[H]
    \centering
    \caption{RAM usage (in MB) for the simulated regression datasets.}
    {\scriptsize 
\begin{tabular}{lllll}
\toprule
Model & \textsf{gse1} & \textsf{gse2} & \textsf{jse2} & \textsf{jse3} \\
\midrule
EigenPro & $440.772 \pm 4.639$ & $493.362 \pm 11.506$ & $457.034 \pm 8.701$ & $454.787 \pm 8.273$ \\
Fastfood & $443.818 \pm 6.483$ & $453.142 \pm 6.608$ & $441.748 \pm 6.308$ & $441.777 \pm 6.313$ \\
GPR & $897.864 \pm 76.904$ & $951.018 \pm 86.28$ & $894.622 \pm 76.644$ & $893.975 \pm 76.439$ \\
Kernel Ridge & $434.895 \pm 6.001$ & $477.738 \pm 9.359$ & $437.968 \pm 4.703$ & $438.134 \pm 4.582$ \\
Nyström & $442.928 \pm 4.844$ & $459.605 \pm 7.743$ & $443.193 \pm 5.298$ & $446.374 \pm 7.932$ \\
\texttt{RFFNet} & $434.291 \pm 7.541$ & $463.314 \pm 10.797$ & $436.339 \pm 4.793$ & $435.628 \pm 4.523$ \\
SRFF & $686.998 \pm 62.039$ & $637.538 \pm 72.795$ & $677.287 \pm 63.376$ & $659.675 \pm 65.885$ \\
\bottomrule
\end{tabular}}
\end{table}

\begin{table}[H]
    \centering
    \caption{RAM usage (in MB) for the simulated classification datasets.}
    {\footnotesize 
\begin{tabular}{lllll}
\toprule
Model & \textsf{classification} & \textsf{moons} \\
\midrule
EigenPro & $447.761 \pm 3.525$ & $437.864 \pm 7.465$ \\
Fastfood & $557.096 \pm 59.894$ & $543.893 \pm 69.235$ \\
Nyström & $529.239 \pm 55.554$ & $524.916 \pm 61.374$ \\
\texttt{RFFNet} & $449.875 \pm 6.953$ & $441.696 \pm 9.223$ \\
\bottomrule
\end{tabular}}
\end{table}

\begin{table}[!ht]
    \centering
    \caption{RAM usage (in MB) for the real-world regression datasets. OOM: out-of-memory.}   
    {\scriptsize
\begin{tabular}{lllll}
\toprule
Model & \textsf{abalone} & \textsf{compact} & \textsf{powerplant} & \textsf{yearprediction} \\
\midrule
EigenPro & $459.742 \pm 9.543$ & $455.401 \pm 16.655$ & $445.329 \pm 4.163$ & OOM \\
Fastfood & $450.48 \pm 13.959$ & $437.146 \pm 7.194$ & $437.583 \pm 1.93$ & $1649.55 \pm 1.488$ \\
GPR & $519.453 \pm 22.149$ & $655.176 \pm 84.886$ & $7985.012 \pm 1194.386$ & OOM \\
Kernel Ridge & $443.39 \pm 5.607$ & $465.184 \pm 18.198$ & $474.495 \pm 8.587$ & OOM \\
Nyström & $428.799 \pm 2.099$ & $438.998 \pm 7.674$ & $434.381 \pm 2.475$ & $1656.081 \pm 4.51$ \\
\texttt{RFFNet} & $429.401 \pm 2.586$ & $432.024 \pm 6.467$ & $455.584 \pm 7.649$ & $1647.918 \pm 6.724$ \\
SRFF & $495.449 \pm 21.55$ & $524.502 \pm 43.78$ & $454.3 \pm 8.655$ & $1662.739 \pm 3.649$ \\
\bottomrule
\end{tabular}}
\end{table}

\begin{table}[H]
    \centering
    \caption{RAM usage (in MB) for the real-world classification datasets. OOM: out-of-memory.}  
    {\scriptsize 
\begin{tabular}{lllll}
\toprule
Model & \textsf{a9a} & \textsf{amazon} & \textsf{higgs} & \textsf{w8a} \\
\midrule
EigenPro & $599.486 \pm 13.495$ & OOM & OOM & $931.943 \pm 0.977$ \\
Fastfood & $805.98 \pm 116.562$ & $8397.65 \pm 337.782$ & $8680.3 \pm 8.591$ & $1210.881 \pm 76.276$ \\
Nyström & $790.729 \pm 125.592$ & $5475.613 \pm 403.35$ & $8680.3 \pm 8.591$ & $1228.162 \pm 161.9$ \\
\texttt{RFFNet} & $599.699 \pm 16.64$ & $4781.993 \pm 8.011$ & $8680.3 \pm 8.591$ & $942.079 \pm 2.453$ \\
\bottomrule
\end{tabular}}
\end{table}

\section{A toy model for \rffnet{}'s landscape}
\label{appendix:convexity}

\begin{figure}[!htb]
\begin{subfigure}[b]{\textwidth}
    \centering
    \includegraphics[scale=0.8]{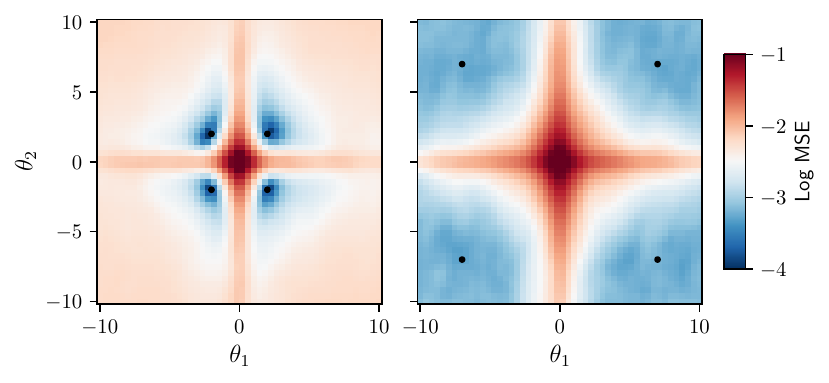}
    \caption{Increasing the \textit{oracle relevance} degrade the identification of relevant features.}  
    \label{fig:convexity1}
\end{subfigure}

\begin{subfigure}[b]{\textwidth}
    \centering
    \includegraphics[scale=0.8]{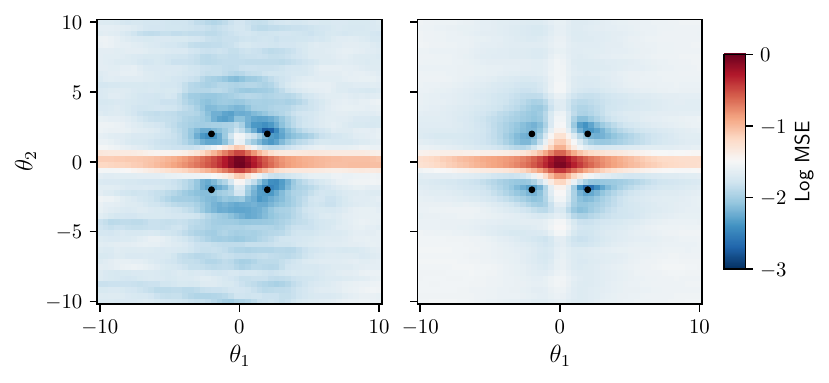}
    \caption{Increasing the sample size improves the identification of relevant features.}
    \label{fig:convexity2}
\end{subfigure}
\end{figure}

In this section, we assess the effect of the sample sizes $n$, the number of random features $s$, and the value of the \textit{population relevance vector} $\theta^*$ (within a restricted data-generating model) in \rffnet{}'s objective. We seek to understand how \rffnet{}, even with a highly non-convex objective function, can identify the relevant features of the data. Our simulations manifest the parity symmetry (with respect to the entries of the relevance vector) in the loss landscape, as well as a type of effective convexity that might emerge in the large-sample setting. 

We fix the number of features as $d = 2$ for visualization purposes only. We consider a ``natural'' input distribution with $\x_i \sim \mathsf{N}(0, I_p)$, $i \in [n]$, where $n$ is the sample size. We work on a completely specified model for the true regression function: if $k_{\theta^*}$ is the ARD Gaussian kernel, with $\theta^* = (\theta_1, \theta_2)$ the population relevance vector, we choose the regression function as
\begin{equation*}
    f^*(\x) =\alpha \, k_{\theta^*}(\x_1, \x),
\end{equation*}
where we sample $\alpha, \x_1 \sim \mathsf{N}(0, 1)$.

We consider the loss landscape with respect to the relevances as the following empirical risk functional:
\begin{equation}
    h_{\beta}(\theta_1, \theta_2) = \frac{1}{n} \sum_{i=1}^n  \ell\left(f^*(\x_i), f_{\theta, \beta}(\x_i) \right),
    \label{eq:landscape}
\end{equation}
where $f^* \in \sH^*$, the RKHS uniquely associated to $k_\theta*$ and $f_{\theta, \beta} \in \sF_{\theta, \beta}$, the \rffnet{}'s hypothesis class, 
\begin{equation*}
    \sF_{\theta, \beta} = \{ f_{\theta, \beta}(\cdot) = \beta^\t \z(\theta \circ \cdot) : \beta \in \R^s, \theta \in \R^d \},
\end{equation*}
with $\z: \R^d \to \R^s$ the random Fourier features map for $k_{\mathbf{1}_d}$. For our simulations, to factor out the influence of $\beta$, we choose it based on the fact that 
\begin{align*}
    f^*(\x) = \alpha \, k_{\theta^*}(\x_1, \x) &\approx \alpha \z(\theta^* \circ \x_1)^\t \z(\theta^* \circ \x) \\ 
    &= \Bigl( \alpha  \z(\theta^* \circ \x_1) \Bigr)^\t \z(\theta^* \circ \x) \\ 
    &\approx \Bigl( \underbrace{ \alpha \z(\theta \circ \x_1)}_{\beta} \Bigr)^\t \z(\theta \circ \x), 
\end{align*}
where the last approximation holds when $\theta \to \theta^*$, and the first when $s$ is sufficiently big. Thus, we take $\beta$ in \eqref{eq:landscape} as
\begin{equation*}
{\beta} = \alpha \, \z(\theta \circ \x_1).
\end{equation*}

In Figure~\ref{fig:convexity1}, we fixed the number of samples as $n = 500$ and the number of random features as $s = 300$. For the left panel, we chose $\theta^* = (2,2)$, while for the right panel, $\theta^* = (7, 7)$. We observe that increasing the relevance vectors degrades the identification of the relevant features, as the loss landscape becomes flat around the possible population relevances. A possible reason is that greater relevances cause functions in $\sH^*$ to oscillate in smaller length scales; as a consequence, samples of the true regression function may be aliased, hindering the detection of the relevant features within the dataset. 

In Figure~\ref{fig:convexity2}, we fixed $\theta^* = (2, 2)$. For the left panel, we choose $(n, s) = (50, 200)$, while for the right panel $(n, s) = (1000, 200)$. We observe that increasing the sample size smooths the loss landscape, damping the ripples created by the oscillatory nature of the random Fourier features map. As a consequence, the loss wells associated with the population relevances become more pronounced. 

In conjunction, these figures indicate that the loss landscape of \rffnet{}, in the large-sample setting with ``low-frequency'' regression functions, might exhibit beneficial properties. We have seen that a simple choice of $\beta$ renders the \rffnet{} objective ``effectively'' convex within each orthant of $\R^d$, with minima that correspond exactly to the population relevance parameter.

Scripts to reproduce these figures are supplied along the code and available in the \texttt{experiments/convexity} folder.

\bibliographystyle{elsarticle-harv} 
\bibliography{bibliography}


\end{document}